\documentclass{article}

\usepackage[final]{neurips_2022}

\usepackage[utf8]{inputenc} 
\usepackage[T1]{fontenc}    
\usepackage{hyperref}       
\usepackage{url}            
\usepackage{booktabs}       
\usepackage{amsfonts}       
\usepackage{nicefrac}       
\usepackage{microtype}      
\usepackage{xcolor}         

\usepackage{graphicx} 
\usepackage{subcaption}

\usepackage{xcolor}
\usepackage{glossaries}
\usepackage{cleveref}
\usepackage{bbm}
\usepackage{amsfonts}
\usepackage{amsthm}
\usepackage{paralist}
\usepackage{cancel}
\usepackage{multirow}
\usepackage{graphicx}
\usepackage{nicefrac}
\usepackage{wrapfig}
\usepackage{algorithm}
\usepackage{algorithmic}

\definecolor{red}{HTML}{E41A1C}
\definecolor{orange}{HTML}{FF7F00}
\definecolor{yellow}{HTML}{FFC020}
\definecolor{green}{HTML}{4DAF4A}
\definecolor{blue}{HTML}{377EB8}
\definecolor{purple}{HTML}{984EA3}

\Crefname{algocf}{Algorithm}{Algorithms}
\crefname{algorithm}{Algorithm}{Algorithms}
\crefname{figure}{Figure}{Figure}
\crefname{table}{Table}{Table}
\crefname{section}{§}{§§}
\Crefname{section}{§}{§§}
\crefformat{equation}{(#2#1#3)}
\crefname{property}{Property}{Property}
\crefname{theorem}{Theorem}{Theorem}
\crefname{proposition}{Proposition}{Proposition}
\crefname{lemma}{Lemma}{Lemma}

\glsdisablehyper{}
\newglossaryentry{lip}
{
    name=lipschitz,
    description={Lipschtiz constant}
}
\newglossaryentry{Lip}
{
    name=lipschitz,
    description={Lipschtiz constant}
}

\newacronym{nn}{NN}{Neural Networks}
\newacronym{sgd}{SGD}{stochastic gradient descent}
\newacronym{svd}{SVD}{Singular Value Decomposition}
\newacronym{sota}{SOTA}{state-of-the-art}
\newacronym{gan}{GAN}{Generative Adversarial Networks}
\newacronym{srn}{SRN}{Stable Rank Normalization}
\newacronym{sngan}{SN-GAN}{Spectral Normalization GAN}
\presetkeys{todonotes}{%
  backgroundcolor=blue!10!white,
  linecolor=blue!10!white,
  bordercolor=blue!10!white
}{}

\DeclareMathOperator{\E}{{}\mathbb{E}}

\newcommand{\SKIP}[1]{}

\newcommand{\bfx}{\mathbf{x}}
\newcommand{\bfp}{\mathbf{p}}
\newcommand{\bfy}{\mathbf{y}}

\newcommand{\real}{\mathbb{R}}

\newcommand{\bfs}{\mathbf{s}}

\newcommand{\vmat}{\mathbf{V}}
\newcommand{\umat}{\mathbf{U}}
\newcommand{\hmat}{\mathbf{H}}

\newcommand{\bfw}{\mathbf{w}}

\newcommand{\dataset}{\mathcal{D}}

\title{RegMixup: Mixup as a Regularizer Can Surprisingly Improve Accuracy \& Out-of-Distribution Robustness}

\author{%
  Francesco Pinto\thanks{FP continued spending time on this project during his internship at Five AI from June 2021 to March 2022.} \\
  University of Oxford \\
  \texttt{francesco.pinto@eng.ox.ac.uk}
  \And
  Harry Yang \\
  Meta AI
  \AND
  Ser-Nam Lim \\
  Meta AI
  \And
  Philip H.S. Torr \\
  University of Oxford
  \And
  Puneet K. Dokania \\
  University of Oxford \& Five AI Ltd. \\
  \texttt{puneet.dokania@five.ai}
}

\begin{document}

\maketitle

\begin{abstract}
 
{\em 
We show that the effectiveness of the well celebrated Mixup~\citep{zhang2018mixup} can be further improved if instead of using it as the sole learning objective, it is utilized as an additional regularizer to the standard cross-entropy loss. This simple change not only improves accuracy but also significantly improves the quality of the predictive uncertainty estimation of Mixup in most cases under various forms of covariate shifts and out-of-distribution detection experiments. In fact, we observe that Mixup otherwise yields much degraded performance on detecting out-of-distribution samples possibly, as we show empirically, due to its tendency to learn models exhibiting high-entropy throughout; making it difficult to differentiate in-distribution samples from out-of-distribution ones. 
To show the efficacy of our approach (RegMixup\footnote{Code available at: \url{https://github.com/FrancescoPinto/RegMixup}}), we provide thorough analyses and experiments on vision datasets (ImageNet \& CIFAR-10/100) and compare it with a suite of recent approaches for reliable uncertainty estimation. 

}
\end{abstract}

\section{Introduction}
In real-world machine learning applications one is interested in obtaining models that can reliably process novel inputs. However, though deep learning models have enabled breakthroughs in multiple fields, they are known to be unreliable when exposed to samples obtained from a distribution that is different from the training distribution. Usually, the larger the extent of this difference between the train and the test distributions, the more unreliable these models are. This observation has led to a growing interest in developing deep learning models that can provide \textit{reliable} predictions even when exposed to unseen situations~\citep{Liu2020-sngp, liu2020energybased,mixupEnsembles, Balaji2017-DeepEns}. Most of these approaches either use expensive ensembles, or propose non-trivial modifications to the neural network architectures in order to obtain reliable models. In most cases, they trade in-distribution performance (accuracy) to perform reliably on: (1) out-of-distribution (OOD) samples; and (2) covariate shift (CS)~\citep{DataShiftBook} samples. 

Towards developing practically useful and reliable models, we investigate the well known Mixup~\citep{zhang2018mixup} as it is extremely popular in improving both a model's accuracy and its robustness~\citep{mixupEnsembles,augmix}. It has already been observed that Mixup can help in retaining good accuracy when the test inputs are affected by superficial variations that do not affect the target label (i.e., they undergo CS) \citep{augmix}. However, we find that Mixup's reliability degrades significantly when exposed to completely unseen samples with potentially different labels than the ones it was shown during training (OOD). This is undesirable as in such situations we would like our models to reliably reject these inputs instead of making wrong predictions on them. 
We observe that the  primary reason for such poor OOD performance of Mixup is due to its tendency to provide high predictive entropy for almost all the samples it receives. 
Therefore, it becomes difficult for the model to differentiate in-distribution samples from out-of-distribution ones. We would like to highlight that our observation is in contrast to the prior work \citep{MixupOOD} which suggests that Mixup provides reliable uncertainty estimates for OOD data as well.

We propose a simple yet effective fix to the aforementioned issue with Mixup: we suggest to train the model on an approximate data-distribution that is an \textit{explicit} mixture of both the  empirical and the Mixup vicinal approximations to the data-distribution. We call this approach RegMixup. In a nutshell, it simply combines the Empirical Risk Minimization (ERM)~\citep{ERM} and the Vicinal Risk Minimization (VRM)~\citep{VRM} objectives together.  Implementation wise, along with the Mixup objective on the interpolated sample, it  requires adding an additional cross-entropy loss on one of the the clean samples. We provide proper justifications behind this proposal and show that such simple modification significantly improves the performance of Mixup on a variety of experiments.

We would like to highlight that one of the core strengths of our approach is its \textbf{simplicity}. As opposed to the recently proposed techniques to improve uncertainty estimation like SNGP~\citep{Liu2020-sngp} and DUQ~\citep{Van_Amersfoort2020-DUQ}, it does not require any modifications to the architecture and is extremely simple to implement. \textit{It also does not trade accuracy in order to improve uncertainty estimates}, and is a single deterministic model, hence, extremely efficient compared to the highly competitive Deep Ensembles (DE) \citep{Balaji2017-DeepEns}. 
Summary of our contributions:
\begin{compactitem}
    \item We provide a simple modification to Mixup that significantly improves its in-distribution, covariate shift, and out-of-distribution performances.
    \item Through extensive experiments using ImageNet-1K, CIFAR10/100 and their various CS counterparts along with multiple OOD datasets we show that, overall, RegMixup outperforms recent state-of-the-art single-model approaches. In most cases, it outperforms the extremely competitive and expensive DE as well.
\end{compactitem}

\section{RegMixup: Mixup as a regularizer}
\paragraph{Preliminary on ERM and VRM} The principle of risk minimization~\citep{ERM} is to estimate a function $f \in \mathcal{F}$ that, for a  given loss function $\ell(.,.)$, minimizes the expected risk over the data-distribution $P(\bfx,\bfy)$. The risk to be optimized is defined as \( R(f) = \int \ell(f(\bfx), \bfy) dP(\bfx,\bfy) \). Since the distribution $P(\bfx,\bfy)$ is unknown, a crude yet widely used approximation is to first obtain a training dataset $\dataset = \{(\bfx_i, \bfy_i)\}_{i=1}^n$ sampled from the distribution $P$ and then obtain $f$ by minimizing the \textit{empirical risk} defined as \( R_{e}(f) = 1/n \sum_{i=1}^n \ell(f(\bfx_i), \bfy_i) \). This is equivalent to approximating the entire data-distribution space by a finite $n$ number of Delta distributions positioned at each  $(\bfx_i, \bfy_i)$, written as $P_e (\bfx, \bfy) = 1/n \sum_{i=1}^n \delta_{\bfx_i}(\bfx)\delta_{\bfy_i}(\bfy)$. This approximation to the risk minimization objective is widely known as the  \textit{Empirical Risk Minimization (ERM)}~\citep{ERM}. 

ERM has been successfully used in a plenty of real-world applications and undoubtedly has provided efficient and accurate solutions to many learning problems. However, it is straightforward to notice that the quality of such ERM solutions would rely on how closely $P_e$ mimics the true distribution $P$, and also on the capacity of the function class $\mathcal{F}$. In situations where the function class is extremely rich with high capacity (for example, deep neural networks), and hence prone to undesirable behaviours such as overfitting and memorization, a good approximation to $P$ is generally needed to enforce suitable inductive biases in the model. 
To this end, for a fixed training dataset, instead of a delta distribution one could potentially fit a richer distribution in the vicinity of each input-output pair to estimate a more informed risk computed in a \textit{region} around each sample. This is precisely the principle behind \textit{Vicinal Risk Minimization (VRM)}~\citep{VRM}.
The approximate distribution in this case can be written as $P_v(\bfx, \bfy) = 1/n \sum_{i=1}^n P_{\bfx_i, \bfy_i}(\bfx, \bfy)$, where $P_{\bfx_i, \bfy_i}(\bfx, \bfy)$ denotes the user-defined vicinal distribution around the $i$-th sample\footnote{Note, the original VRM paper uses $P_{\bfy_i}(\bfy) = \delta_{\bfy_i}(\bfy)$ which simply is a special case of this notation.}. 
Therefore, the \textit{vicinal risk} boils down to

\begin{align}
\label{eq:vrm-risk}
    R_v(f) = \frac{1}{n}\sum_{i=1}^n \int \ell(f(\bfx), \bfy) dP_{\bfx_i, \bfy_i}(\bfx, \bfy).
\end{align}

When the integral in the above summation is  intractable, a Monte Carlo estimate with $m$ samples can be used as follows:
\begin{align}
\label{eq:mc-risk}
    \int \ell(f(\bfx), \bfy) dP_{\bfx_i, \bfy_i}(\bfx, \bfy) \approx \frac{1}{m}\sum_{j=1}^m \ell(f(\bar{\bfx}_j), \bar{\bfy}_j); \; \; (\bar{\bfx}_j, \bar{\bfy}_j) \sim P_{\bfx_i, \bfy_i}(\bfx, \bfy).
\end{align}
Several approaches in deep learning can be seen as a special instance of VRM. For example, training a neural network with multiple augmentations is a special case where the augmented inputs are the samples from the unknown vicinal distribution. 
A widely used application of VRM is the procedure to obtain a  \textit{robust base classifier} to design certifiable classifiers. 
\begin{wrapfigure}{r}{0.4\textwidth}
    \centering
    \includegraphics[width=0.32\textwidth]{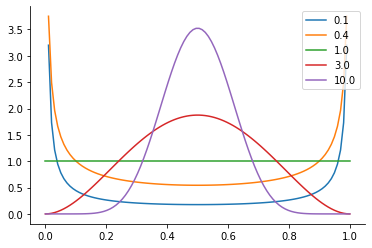}
    \caption{$\texttt{Beta}(\alpha,\alpha)$ pdf for varying $\alpha$.}
    \label{fig:betas}
\end{wrapfigure}
For example, if $P_{\bfx_i, \bfy_i}(\bfx, \bfy) = P_{\bfx_i}(\bfx) \delta_{\bfy_i}(\bfy)$ and $P_{\bfx_i}(\bfx)$ is a Gaussian distribution centered at $\bfx_i$, Eq.~\eqref{eq:mc-risk} can be computed by taking the average loss over gaussian perturbed inputs $\bar{\bfx}$ while keeping the target labels same. Minimizing such a risk would lead to a classifier that is robust to additive noise bounded within an $\ell_2$ ball. This is exactly the procedure that has been widely adopted in the \textit{randomized smoothing} literature in order to obtain a base neural network for which a \textit{certifiable} smooth classifier can be obtained~\citep{lecuyer2019certified,cohen2019certified}\footnote{Based on our understanding, the current literature do not mention this procedure as an instance of VRM.}. Below we discuss another highly effective use case of VRM called {\em Mixup} which is the main focus of this work.

\paragraph{Mixup} Built on the fundamentals of VRM, the vicinal distribution defined in Mixup~\citep{zhang2018mixup} is as follows:  
\begin{align}
   P_{\bfx_i, \bfy_i}(\bfx, \bfy) =  \E_{\lambda}  [ (\delta_{\bar{\bfx}_i}(\bfx), \delta_{\bar{\bfy}_i}(\bfy)) ], \nonumber
\end{align}
where $\lambda \sim \texttt{Beta}(\alpha,\alpha) \in [0, 1]$, $\alpha \in (0,\infty)$, $\bar{\bfx}_i = \lambda \bfx_i + (1-\lambda) \bfx_j$ and $\bar{\bfy}_i = \lambda \bfy_i + (1-\lambda) \bfy_j$. Note that the vicinal  distribution here not only depends on $(\bfx_i, \bfy_i)$ but also on another input-output pair $(\bfx_j, \bfy_j)$ drawn from the same training dataset. For a fixed $\alpha$ (parameter of the Beta distribution, refer Figure~\ref{fig:betas}), implementing Mixup would require taking multiple Monte Carlo samples\footnote{We do not investigate how $m$ depends on $\alpha$.}
for each datapoint (refer Eq.~\eqref{eq:mc-risk}) which can be computationally prohibitive. Therefore, in practice, only one sample ($m=1$) per Beta distribution per pair of samples from a batch is considered at a time. Although this procedure might look like a crude approximation to the original objective, it has resulted in highly promising results in a variety of applications and is very well accepted in the research community.  Without undermining the remarkable success of such a simple approach, below we highlight two of its behavioural characteristics that, as we show, might limit its effectiveness:
\begin{compactitem}
\item \textbf{Small cross-validated $\alpha \ll 1$:} The shape of the vicinal distribution depends on the hyperparameter of the Beta distribution, therefore, the values of $\alpha$ decides the strength of the convex interpolation factor $\lambda$. 
Since high values of $\alpha$ would encourage $\lambda \approx 0.5$ resulting in an interpolated $\bar{\bfx}$ that is perceptually different from $\bfx$ (\textit{inducing a mismatch between train and test distributions}), the cross-validated value of $\alpha$ for Mixup in most cases turns out to be very small ($\alpha \approx 0.2$) in order to provide good generalization. Such small values of $\alpha$ leads to sharp peaks at $0$ and $1$ (refer Figure~\ref{fig:betas}). Therefore, effectively, Mixup ends up \textit{slightly} perturbing a clean sample in the direction of another sample even though the vicinal distribution has the potential to explore a much larger interpolation space.
\item \textbf{High-entropy behaviour:} As mentioned earlier, $m=1$ is used in practice, therefore, it is very unlikely that the interpolation factor $\lambda \in \{0,1\}$ even for small values of $\alpha$. Thus, the model never gets exposed to uninterpolated clean samples during training and hence, it always learns to predict interpolated (or smoothed) labels $\bar{\bfy}$ for every input. Just like DNNs with cross-entropy loss are overconfident because of their high capacity and target Delta distribution~\citep{GuoCalibration}, DNNs with Mixup turns out to be \textit{relatively less confident} because the network retains its high capacity but observes only target smoothed labels. 
This underconfident behaviour results in high-entropy for both in-distribution and out-of-distribution samples. This is undesirable as it will not allow predictive uncertainty to reliably differentiate in-distribution samples from out-of-distribution ones. 
\end{compactitem}

\begin{figure}[t]
    \centering
    \includegraphics[width=0.8\textwidth]{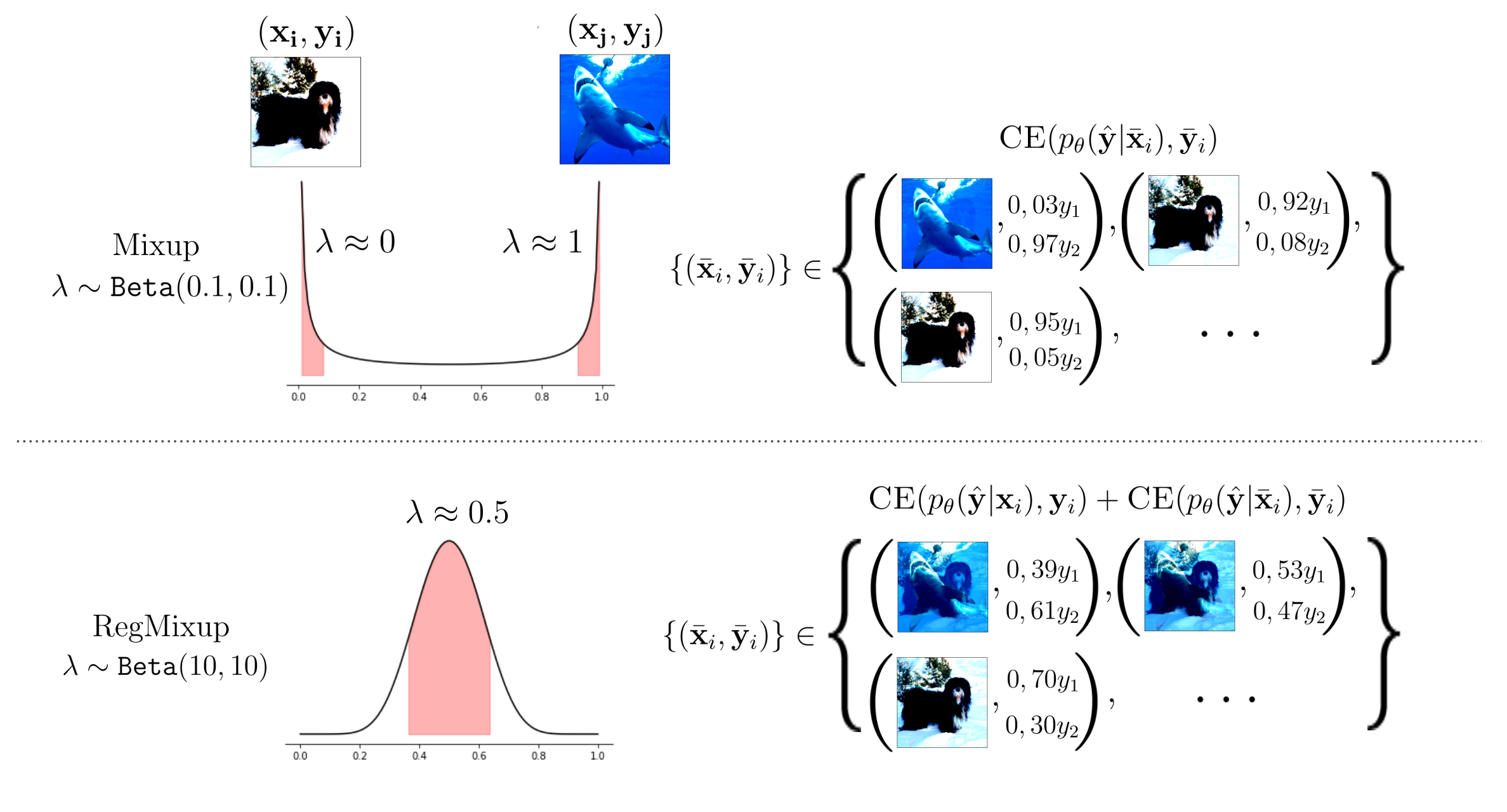}
    \caption{\textbf{Mixup vs RegMixup in practice}. Illustration of how the cross-validated $\alpha$ affects the shape of the $\texttt{Beta}$ distribution in both cases. \textcolor{red}{Red} regions represent 80\% of the probability mass. Mixup typically samples $\lambda \approx 0$ or $1$, while for RegMixup $\lambda \approx 0.5$. In the first case, one of the two interpolating images dominates the interpolated one; in the latter, a wide variety of images containing features from both the images in the pair are obtained. 
    }
    \label{fig:diagram}
\end{figure}

\begin{wrapfigure}{r}{0.4\textwidth}
    \centering
    \includegraphics[width=0.37\textwidth]{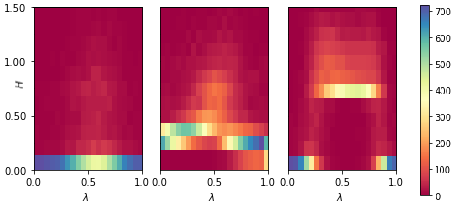}
    \caption{\textbf{Heatmaps of the entropy profiles} as the interpolation factor $\lambda \in [0,1]$ between samples of two classes varies. Left (\textbf{DNN}), Middle (\textbf{Mixup}), Right (\textbf{RegMixup}). Note, RegMixup induces high-entropy barrier separating in-distribution \& out-distribution samples.}
    \label{fig:entropy_profiles}
\end{wrapfigure}

We validate the consequences of the above limitations of Mixup with the following simple experiment. In Figure~\ref{fig:entropy_profiles}, we provide heat-maps to show how the entropy of the predictive distribution varies when interpolating samples belonging to different classes. The heat-map is created as follows. First, we train a WideResNet28-10 (WRN)~\citep{WideResNet} using the CIFAR-10 (C10) dataset. Then, we randomly choose 1K pairs of samples $\{\bfx_i, \bfx_j\}$ from the dataset such that $y_i \neq y_j$\footnote{Note, it is highly \textit{likely} that $\bfy_i \neq \bfy_j$ even if we do not impose this constraint as the problems under consideration have many classes.}. For each pair, via convex combination, we synthesize 20 samples $\bar{\bfx}$s using equally spaced $\lambda$s between $0$ to $1$. 
The heat-map is then created using all the 20K samples. The intensity of each ($\lambda$, H) bin in the heat-map indicates the number of samples in that bin. Note, DNN (trained with vanilla cross-entropy loss) shows low entropy (thus yielding overconfidence) irrespective of where the interpolated sample lies. However, Mixup shows higher entropy for almost all $\lambda$s (thus yielding lower confidence on in-distribution samples as well). The impact of this behaviour is shown in Table~\ref{tab:small_table}. Though Mixup provides improved accuracy compared to DNN for in-distribution (IND) and covariate shift experiments, the high entropy behaviour makes it much worse than DNN when exposed to out-of-distribution detection task. For example, when SVHN is used as the OOD dataset, the performance of Mixup drops by nearly $8.47\%$ compared to DNN. This clearly shows that the distributions of the per-sample entropies on the in-distribution and out-of-distribution sets are harder to separate. However, there is a clear improvement of nearly $5\%$ for covariate shift experiments, implying that Mixup augmentations do improve robustness in this aspect. Note that, in the context of calibration, Mixup's confidence reducing behaviour (equivalent to providing higher predictive entropy) was also observed by~\citep{mixupEnsembles}.

\paragraph{RegMixup} We now provide a very simple modification to Mixup that not only alleviates its aforementioned limitation on detecting OOD samples, but also significantly improves its performance on in-distribution and covariate-shift samples. We propose to use the following approximation to the data-distribution
\begin{align}
    P(\bfx, \bfy) = \frac{1}{n} \sum_{i=1}^n (\gamma \delta_{\bfx_i}(\bfx)\delta_{\bfy_i}(\bfy) + (1- \gamma) P_{\bfx_i, \bfy_i}(\bfx, \bfy)), \nonumber
\end{align}
where $P_{\bfx_i, \bfy_i}(\bfx, \bfy)$ is the mixup vicinal distribution and $\gamma \in [0, 1]$ is the mixture weight. \textit{The above approximation is simply an explicit assemble of ERM and VRM} based approximations to data-distribution. Though, in theory, VRM subsumes ERM, we argue that because of the  
crude approximations needed to prevent increasing the training cost, it does not fully realize its potential. Therefore, explicitly combining them might result in a \textit{practically} more expressive model.  
We provide extensive experimental evidence to support this hypothesis. Implementation wise, for each sample $\bfx_i$ in a batch, another sample $\bfx_j$ is drawn randomly from the same batch to obtain interpolated $\bar{\bfx}_i$ and $\bar{\bfy}_i$, and then the following loss is minimized
\begin{align}
    \textsc{CE} (p_{\theta}(\hat{\bfy} | \bfx_i), \bfy_i) + \eta \; \textsc{CE} (p_{\theta}(\hat{\bfy} | \bar{\bfx}_i), \bar{\bfy}_i), 
    \label{eq:RegMixup}
\end{align}
where $\textsc{CE(.,.)}$ denotes the standard cross-entropy loss, the hyperparameter $\eta \in \real_{\geq 0}$, and $p(.)$ the softmax output of a neural network parameterized by $\theta$. Note, dividing Eq.~\eqref{eq:RegMixup} by $(1+\eta)$ is exactly the same as using a data-distribution with $\gamma = \nicefrac{1}{1+\eta}$. In practice, we observe the model's performance to not vary much with $\eta$ and using $\eta = 1$ (equivalent to $\gamma = 0.5$) to provide highly effective results. Refer to Algorithm~\ref{alg:RegMixup} for an overview of the RegMixup training procedure.

\textit{What does this simple modification bring to the model?} (1) VRM now subsumes ERM because the vicinal distribution can perfectly represent the ERM training distribution.
This is because of the fact that irrespective of the values of $m$ and $\alpha$, the model will always be exposed to the clean training samples as well. (2) The interpolation factor $\lambda$ can potentially explore a much wider space as the presence of clean samples might help in controlling the performance drop due to the train/test distribution shift. Therefore, if $\alpha \ll 1$ was actually the most effective solution, the cross validation would automatically find it.   

\textit{Practical implications of such simple modification on the behaviour and the performance of the model}
\begin{compactitem}
\item \textbf{Large cross-validated $\alpha \gg 1:$} As anticipated, the model is now able to explore strong interpolations because of the additional cross-entropy term over the unperturbed training data. Interestingly, the cross-validated $\alpha$ that we obtained in fact is very high ($\alpha \in \{10,20\}$, see Appendix \ref{sec:cv} for cross-validation details) leading to $\lambda \approx 0.5$. Therefore, as opposed to the standard Mixup, RegMixup prefers having \textit{strong diverse interpolations} during training. Refer Figure~\ref{fig:diagram} for visualizations. 
\item \textbf{Maximizing a soft proxy to entropy (knowingly unknowing the unknown):} It is straightforward to notice that a value of $\lambda \approx 0.5$ would lead to $\bar{\bfx}$ that is a heavy mix of two samples (mimicking OOD samples, refer Fig.~\ref{fig:diagram}), and the corresponding target label vector $\bar{\bfy}$ would have almost equal masses corresponding to the labels of the interpolating samples. Therefore, minimizing \( \textsc{CE} (p_{\theta}({\bfy} | \bar{\bfx}_i), \bar{\bfy}_i) \) in this case is equivalent to \textit{maximizing a soft proxy to entropy} defined over the label support of $\bfy_i$ and $\bfy_j$ (note, exact entropy maximization would encourage equal probability masses of 0.5 for these labels). We find this observation intriguing as RegMixup naturally obtains a cross-validated $\alpha$ that leads to a maximum likelihood solution having high-entropy on heavily interpolated samples.
This is an extremely desirable property as it allows models to reliably differentiate between in and out distribution samples. Thus, Mixup automatically acts as a \textbf{regularizer} in this case. 
\end{compactitem}

\begin{wraptable}{r}
{0.6\textwidth}
\centering
\resizebox{\linewidth}{!}
{
	\begin{tabular}{cc|c|ccc}
		\toprule
& \multicolumn{1}{c}{} & \multicolumn{1}{c}{\textbf{Cov. Shift}} & 
		\multicolumn{3}{c}{\textbf{OOD Detection}} \\
		&\multicolumn{1}{c}{\textbf{C10} (Test)}
		&\multicolumn{1}{c}{\textbf{C10-C}} & 
		\multicolumn{1}{c}{\textbf{C100}} &
		\multicolumn{1}{c}{\textbf{SVHN}} &
		\multicolumn{1}{c}{\textbf{T-ImageNet}} \\
		\textbf{Models} & \textbf{Accuracy} ($\uparrow$) &
		  \textbf{Accuracy} ($\uparrow$) &
		  \textbf{AUROC} ($\uparrow$) &
		  \textbf{AUROC} ($\uparrow$) &
		  \textbf{AUROC} ($\uparrow$) \\
		\midrule

DNN             &   96.14  & 76.60 &  88.61  & 96.00 & 86.44   \\
Mixup           &  97.01  &81.68 &  83.17 & 87.53 & 84.02         \\
RegMixup (Our)      &  \textbf{97.46} &\textbf{83.13 } & \textbf{89.63} & \textbf{96.72} & \textbf{90.19 } \\

\bottomrule 
\end{tabular}

}

    \caption{In-distribution, covariate shift, and out-of-distribution detection results for WRN trained on CIFAR-10 (C10). C10-C is the corrupted version of C10. }
    \label{tab:small_table}
\end{wraptable}

The entropy heat-map in Figure~\ref{fig:entropy_profiles} clearly shows that as opposed to DNN and Mixup, the entropy for RegMixup is very low for $\lambda$ close to either $0$ or $1$, however, it increases and remains high for all other intermediate interpolation factors, \textit{practically creating an entropy barrier}. 

Also, as shown in Table~\ref{tab:small_table}, RegMixup provides improvements in both accuracy (in-distribution and covariate-shift) and out-of-distribution robustness. An interesting observation is that Mixup's performance on OOD (T-ImageNet) dropped by 2.4\% compared to DNN whereas RegMixup performed 3.75\% better than DNN. Thus, in this case, it effectively improved the performance of Mixup by nearly 6.15\%.

\section{Related Works}  

\paragraph{Covariate-Shift Robustness} It is well known that neural networks performance can severely degrade when exposed to covariate shift \citep{DataShiftWrong,ImageNetv2, ImageNet-Sk}. In image classification, for instance, covariate shift could occur due to changes in environmental conditions \citep{hendrycks2019cifarc} and the image capturing process, data sampling  \citep{ImageNetv2, Cifar10.1, Cifar10.2}, stylised representations of the objects (e.g. in paintings or drawings) \citep{ImageNet-Sk, ImageNetR} etc.
In the field of out-of-distribution generalization, many techniques assume the availability of sets of samples presenting such variations at training time \citep{shi2021gradient, yao2022improving,wang2021generalizing}, other techniques suggest leveraging augmentations to mimic a set of variations that would allow the model to learn features that generalize better \citep{augmix, TheManyFacesOfRobustnessDeepMix} or self-supervision \citep{selfsupervisionhendrycks, contrastiveConnects}.

\paragraph{Out-Of-Distribution Detection} While some authors suggest introducing an OOD reject option \citep{geifman2019selectivenet}, the most recent literature tends to focus on leveraging a threshold classifier on an uncertainty measure computed on top of the predictive distribution of a network. For the latter approach, it is therefore convenient for a model to produce high uncertainty predictions when facing OOD samples, and low uncertainty predictions when fed in-distribution samples \citep{Liu2020-sngp}. Within this family of approaches, the most popular ones are the expensive  Deep Ensembles (DE)~\citep{Balaji2017-DeepEns}. Note, the computational cost of DE scales linearly with the number of members, and also recent works have observed a few limitations of naively using ensembles directly~\citep{mixupEnsembles, pitfallsEnsemble}. There are several efficient variants of DE where the ensemble members are obtained either by processing multiple inputs and multiple outputs at the same time~\citep{havasi2021-MIMO}, adding member-specific components to the network~\citep{Wen2020-BatchEnsemble}, or by finding optima in distinct basins of low loss~\citep{snapshotEnsembles}. Significant attention has also been paid to Bayesian techniques where either stochastic gradient descent is being modified to perform approximate inference~\citep{sotchGradHMC,CyclicalStochMCMC,StochGRRMCMC} or an approximation to Gaussian Processes is being utilized (specifically, we focus on SNGP \citep{Liu2020-sngp}), or Bayesian Logistic Regression is being used during inference via Laplace approximation (specifically, we focus on KFAC-LLLA ~\citep{HeinBeingBayesianABit2020}). Techniques utilizing distance functions in the feature space have also been suggested \citep{Van_Amersfoort2020-DUQ}.   

\paragraph{Model calibration} Modern NNs have been shown to be miscalibrated~\citep{GuoCalibration}, i.e. they exhibit a mismatch between the model's confidence and its accuracy. It has been observed that temperature scaling~\citep{GuoCalibration} or replacing the typical cross-entropy loss \citep{beyondpinball, MixupOOD, mukhoti2020calibrating} can be highly effective to reduce this mismatch. 
Also, ensemble learning has been observed to help in reducing miscalibration \citep{Balaji2017-DeepEns, mixupEnsembles, uncertaintyEnsemblesTScaling}. However, it has been shown that good calibration on in-distribution data does not necessarily imply good calibration performance under covariate-shift~\citep{ovadia2019trust}.

\section{Experiments}
\label{sec:experiments}
\paragraph{Datasets and Network Architectures} We employ the widely used WideResNet28-10 (WRN) ~\citep{WideResNet} and ResNet50 (RN50) ~\citep{ResNet} architectures. We train them on CIFAR-10 (C10) and CIFAR-100 (C100) datasets. We employ RN50 to perform experiments on ImageNet-1K \citep{Imagenet1K} dataset. We report the average of all the metrics computed on 5 seeds. For further details about the code base and the hyperparameters, refer to Appendix \ref{sec:experimentDetails}.

For \textbf{Covariate Shift (CS)} experiments on models trained on C10 and C100, we resort to the widely used CIFAR10-C (C10-C) and CIFAR100-C (C100-C) datasets, corrupted versions of C10 and C100~\citep{hendrycks2019cifarc}. These datasets are made by applying $15$ synthetically generated but \textit{realistic corruptions} at $5$ degrees of intensity on the test sets of C10 and C100, respectively. For CIFAR-10, we also use the CIFAR-10.1 (C10.1)~\citep{Cifar10.1} and CIFAR-10.2 (C10.2)~\citep{Cifar10.2} datasets designed to test the generalization of CIFAR-10 classifiers to \textit{natural covariate shifts}. To the best of our knowledge there is no such analogous dataset for CIFAR-100.  For ImageNet-1K experiments, we use the widely considered ImageNet-A (A) \citep{NaturalAdversarialExamples}, ImageNet-R (R) \citep{ImageNetR}, ImageNetv2 (V2) \citep{ImageNetv2}, and  ImageNet-Sketch (SK) \citep{ImageNet-Sk} datasets for covariate shift experiments. We report the calibration experiments on both in-distribution and covariate shifted inputs in Appendix \ref{sec:miscalibr}.

For \textbf{Out-of-Distribution (OOD)} detection, following SNGP~\citep{Liu2020-sngp}, we use C100 and SVHN~\cite{SVHN} as OOD datasets for models trained on C10. Similarly, for models trained on C100, we use C10 and SVHN as OOD datasets. Additionally, we also consider the Tiny-ImageNet (T-ImageNet) dataset~\citep{TinyImageNet} as OOD set in both the cases. 
For models trained on ImageNet-1K, we use  ImageNet-O (O)  \citep{NaturalAdversarialExamples} as the OOD dataset. 

\paragraph{Methods considered for comparisons} Besides the natural comparison of our method with Mixup~\citep{zhang2018mixup} and networks trained via ERM on the standard cross-entropy loss (which we will refer to as DNN), we consider several other methods from the OOD detection and CS literature. For models trained on C10 and C100, we consider:
\begin{compactitem}
    \item DNN-SN and DNN-SRN: taking inspiration from~\citep{Liu2020-sngp}, we consider DNN models trained with 
    Spectral Normalization (SN)~\citep{Miyato2018-SN} and Stable Rank Normalization (SRN)~\citep{Sanyal2020-SRN} to control the bi-Lipschitz constant of the networks, which has been shown to affect the generalization properties of neural networks.
    \item SNGP: Spectrally Normalized Gaussian Process ~\citep{Liu2020-sngp}.
    \item DUQ: Deterministic Uncertainty Quantification  \citep{Van_Amersfoort2020-DUQ}. 
    \item KFAC-LLLA:  KFAC-Laplace Last Layer Approximation~\citep{HeinBeingBayesianABit2020} makes a model Bayesian at test time by taking the Laplace approximation of the last layer~\citep{Ritter2018aLaplace}. We provide a simple outline of this approach in Appendix~\ref{sec:BayesianTest}.
    \item AugMix~\citep{augmix}: A data augmentation technique that applies randomized augmentations to an input while enforcing them to be consistent during training. With the  recommended hyperparameters in the paper, it is almost 4$\times$ slower than DNN during training while having the same inference requirements.
    \item DE: Deep Ensembles \citep{Balaji2017-DeepEns} with 5 members, requiring 5$\times$ more compute than most single-model approaches such as DNN.
\end{compactitem}
We would like to mention that, compared to vanilla DNN, our approach (RegMixup) is almost 1.5$\times$ slower and Mixup is about 1.2$\times$ slower during training, while having the same inference requirements. 
Due to the high compute requirements, for ImageNet-1K we consider DNN, Mixup and the two other strongest baselines: AugMix and DE. We also cross-validate the hyperparameters on a $10\%$ split of the test set, which is removed at test time.

\paragraph{Few missing experiments:} Below we provide extensive experiments for proper benchmarking. The number of datasets and architectures we use lead to many combinations, a few of which we were not able to produce good results for (even after extensive hyperparameter search), hence, some of the entries in the tables are missing. For example, we could not make DUQ work on C100 as it exhibited unstable behaviour. We could not produce promising results for SNGP on RN50 CIFAR experiments using their official implementation.  Similarly for AugMix RN50 experiments on C10 and C100. We chose not to report these suboptimal numbers. Further details can be found in Appendix~\ref{sec:experimentDetails}.

\paragraph{Table entries:} Bold represents the best among all the single-model approaches, and underlined represents the best among all including the expensive Deep Ensembles. 

Note, we do not consider methodologies requiring access to an external dataset during training (either for CS or OOD ) as not only this would be an unfair comparison\footnote{since all the methods we consider do not leverage this information.  RegMixup only relies on in-distribution training data, just like other approaches, and makes no assumption on the type of CS nor on the OOD inputs.}, but we believe assuming such knowledge is against the goal of this work, which is to develop models robust to unknown scenarios. 
\subsection{RegMixup Improves Accuracy on In-distribution and Covariate-shift Samples}

\paragraph{Small-scale (CIFAR) Experiments on In-distribution Data} In Table \ref{tab:new_acc_wrn}, we compare the accuracy of various approaches on the in-distribution test sets of C10 and C100, respectively.  

\begin{wraptable}{r}{0.5\textwidth}
\vspace{-1.5em}
\resizebox{\linewidth}{!}
{
	\begin{tabular}{ccc|cc}
		\toprule
		& \multicolumn{4}{c}{\textbf{IND Accuracy }} \\

& \multicolumn{2}{c|}{\textbf{WRN28-10}} & \multicolumn{2}{c}{\textbf{RN50}} \\
		& \multicolumn{1}{c}{\textbf{C10 (Test)}} & \multicolumn{1}{c|}{\textbf{C100 (Test)}} & \multicolumn{1}{c}{\textbf{C10 (Test)}} & \multicolumn{1}{c}{\textbf{C100 (Test)}}  \\

		\textbf{Methods} & \multicolumn{2}{c|}{\textbf{Accuracy ($\uparrow$)}} & \multicolumn{2}{c}{\textbf{Accuracy ($\uparrow$)}} \\
		\midrule

DNN             &   96.14  & 81.58 &  95.19 & 79.19    \\
Mixup           &  97.01 & 82.60 &  96.05 & 80.12  \\
RegMixup (\textbf{our})      &  \textbf{\underline{97.46}}  & \textbf{83.25} &  \textbf{\underline{96.71}}   &  \textbf{81.52}   \\
\cmidrule{2-5}
DNN-SN          &   96.22  & 81.60 &  95.20  & 79.27  \\
DNN-SRN         &   96.22  & 81.38 &  95.39  & 78.96 \\
SNGP            & 95.98 & 79.20 &- &- \\
DUQ             & 94.7  &- &- &-  \\  
KFAC-LLLA       & 96.11  & 81.56 & 95.21   & 79.41   \\
Augmix          & 96.40 & 81.10  &- &-  \\ 
\midrule
DE ($5\times$)  & 96.75 & \underline{83.85} & 96.23 & \underline{82.09}  \\
\bottomrule 
\end{tabular}%


}
\caption{Accuracies (\%) on IND samples for models trained on C10 and C100}
\vspace{-2em}
	\label{tab:new_acc_wrn}
\end{wraptable}

Clearly, 
\begin{compactitem}
    \item RegMixup outperforms Mixup in all these experiments. In fact, RegMixup is the best performing one among all the single-model approaches. 
    \item These improvements are non-trivial. For instance, on WRN trained on C100, it outperforms DNN and Mixup by $1.67\%$ and $0.65\%$, respectively. It outperforms SNGP with a significant margin of \textbf{$4.05\%$}.
\end{compactitem}

Note how the single-model approaches specifically designed to provide reliable predictive uncertainty estimations (for example, SNGP, DUQ, KFAC-LLLA) underperform even compared to the vanilla DNN in terms of in-distribution accuracy. In order to provide improved uncertainty estimates (as we will soon show), they trade clean data accuracy. This type of behaviour is not observed in RegMixup.

\paragraph{Small-scale (CIFAR) Covariate Shift Experiments}

 For C10-C and C100-C, as typical in the literature, we report the accuracy averaged over all the corruptions and degrees of intensities in Table \ref{tab:new_dist_shift_wrn}. It is evident that our approach produces a remarkable improvement in the average accuracy compared to \textbf{all} the baselines (except AugMix, we discuss later why that might be the case). For instance, for C100-C WRN experiments, our method achieves an accuracy improvement of \textbf{6.9\%} over DNN, of \textbf{3.86\%} over DE, and of \textbf{2.45\%} over Mixup. Similarly, for C10-C WRN, our method achieves an improvement of almost \textbf{6.53\%} over DNN, of \textbf{4.81\%} over DE, and of \textbf{1.45\%} over Mixup. 
 
 For natural covariate shift datasets C10.1 and C10.2 as well, RegMixup outperforms \textbf{all} the baselines (including AugMix). For instance, on C10.2, it obtains an improvement of \textbf{3.26\%} over DNN, of \textbf{1.5\%} over Mixup, and of \textbf{2.46\%} over the expensive DE. 

\begin{wraptable}{r}{0.5\textwidth}
\vspace{-1.2em}

\resizebox{\linewidth}{!}
{

	\begin{tabular}{ccccc|cccc}

		\toprule
		& \multicolumn{8}{c}{\textbf{Covariate Shift Accuracy }} \\
& \multicolumn{4}{c|}{\textbf{WRN28-10}} &  \multicolumn{4}{c}{\textbf{R50}} \\ 
		& \multicolumn{1}{c}{\textbf{C10-C}} &  \multicolumn{1}{c}{\textbf{C10.1}} &  \multicolumn{1}{c}{\textbf{C10.2}} &  \multicolumn{1}{c|}{\textbf{C100-C}} & \multicolumn{1}{c}{\textbf{C10-C}} &  \multicolumn{1}{c}{\textbf{C10.1}} &  \multicolumn{1}{c}{\textbf{C10.2}} &  \multicolumn{1}{c}{\textbf{C100-C}} \\

		\textbf{Methods} & \multicolumn{4}{c|}{\textbf{Accuracy ($\uparrow$)}} & \multicolumn{4}{c}{\textbf{Accuracy ($\uparrow$)}}  \\
		\midrule
DNN              & 76.60 & 90.73 & 84.79                                                &  52.54 & 75.18 & 88.58 & 83.31 & 50.62 \\
Mixup            &81.68 & 91.29 & 86.55 &56.99 &  78.63 & 90.03 & 84.61 & 53.96     \\
RegMixup (\textbf{our})      & 83.13    & \textbf{\underline{92.79}  }& \textbf{\underline{88.05} } & 59.44  & \textbf{\underline{81.18}}  & \textbf{\underline{91.58}} & \textbf{\underline{86.72}} & \textbf{\underline{57.64}} \\
\cmidrule{2-9}

DNN-SN           &  76.56 & 91.01 & 84.72 &   52.61  & 74.88 & 88.26 & 82.96 & 50.55  \\
DNN-SRN          &  77.21 & 90.88 & 85.24                                              & 52.54 & 75.40  & 88.61 & 83.49 & 50.48 \\
SNGP             & 78.37 & 90.80 & 84.95 & 57.23  & -  & -  & -  & -    \\
DUQ              & 71.6  &-&-& 50.4  & -  & -  & -  & -\\
KFAC-LLLA        &76.56  & 90.68 & 84.68                                               &52.57 & 75.18  & 88.34  & 83.50  &50.85   \\
AugMix & \textbf{\underline{90.02}}   & 91.6 & 85.9 & \textbf{\underline{68.15}}  & -  & -  & -  & -  \\
\midrule 
DE ($\times$ 5)   & 78.32 & 92.17 & 85.59                                                & 55.58  & 77.63  & 90.05 & 85.00 &  53.91      \\

\bottomrule
\end{tabular}%

}
\caption{Accuracies (\%) on covariate shifted samples for models trained on C10 and C100.}
	\label{tab:new_dist_shift_wrn}

\resizebox{0.5\columnwidth}{!}{

\begin{tabular}{cc|cccc|c}
	\toprule
	  & \multicolumn{1}{c}{\textbf{IND Acc.}}& \multicolumn{4}{c|}{\textbf{Covariate Shift Acc}} & \multicolumn{1}{c}{\textbf{OOD Det.}}  \\

	 & \multicolumn{1}{c|}{\textbf{ImageNet-1K}} & \multicolumn{1}{c}{\textbf{R}} &
	\multicolumn{1}{c}{\textbf{A}} &
	\multicolumn{1}{c}{\textbf{V2}} &
	 \multicolumn{1}{c|}{\textbf{Sk}} &
	 \multicolumn{1}{c}{\textbf{O}}  \\

	  & \textbf{Acc} ($\uparrow$) & \textbf{Acc} ($\uparrow$)   &
	  \textbf{Acc} ($\uparrow$) & \textbf{Acc} ($\uparrow$)  & \textbf{Acc} ($\uparrow$) &  \textbf{AUROC} ($\uparrow$)   \\

\midrule 
DNN & 76.60 & 36.41  & 2.76 & 64.53 & 24.72 & 55.97\\ 
Mixup & 77.15 & 39.05 & 3.29 & 64.58 & 26.34 & 55.54\\ 
RegMixup (\textbf{our}) & \textbf{77.68} & \textbf{\underline{39.76}} & \textbf{\underline{5.96}} & \textbf{65.66} & \textbf{26.98} & \textbf{\underline{57.05}}\\ 
\cmidrule{2-7}

AugMix & 76.88 & 38.29 & 2.63 & 64.94 & 25.61 & 56.91\\ 
\midrule 
DE (5$\times$) & \underline{78.22} & 38.94 & 2.11 & \underline{66.68} & \underline{27.03} & 53.29 \\
\midrule 

	\end{tabular}%
}
    \caption{ImageNet accuracies (\%) on IND and CS samples, and OOD detection performance. 
}
    \label{tab:imagenet}
    \vspace{-3em}
\end{wraptable}

\textit{Why AugMix performs extraordinarily well on synthetically corrupted C10-C and C100-C but not on natural distribution shift C10.1 and C10.2?} Looking at Table  \ref{tab:new_dist_shift_wrn} one can observe AugMix's extremely good performance on the C10-C and C100-C. However, at the same time, the model is underperforming with respect to RegMixup on C10.1 and C10.2. Similarly, AugMix is outperformed by RegMixup on all ImageNet CS experiments (and OOD as will be shown soon). This seems to suggest that although the augmentations used during the training of AugMix are not exactly same as that of the corrupted test dataset, they tend to benefit from synthetic forms of covariate shifts, hence the dramatic improvement in these particular scenarios. 

\paragraph{Large-scale (ImageNet-1K) Experiments on In-distribution Data} 
As shown in Table \ref{tab:imagenet}, RegMixup scales to ImageNet-1K and exhibits improved accuracy with respect to both Mixup, DNN, and AugMix. In particular, it is \textbf{1.08\%} better than DNN, \textbf{0.53\%} better than Mixup and \textbf{0.80\%} better than AugMix. DE in this case is the best performing one.

\paragraph{Large-scale (ImageNet-1K) Covariate Shift Experiments} In Table \ref{tab:imagenet} we report the results for common ImageNet-1K covariate-shift test sets. As it can be seen, RegMixup is either the best performing one among all the single-model approaches, or it is the absolute winner including DE. For instance, on ImageNet-A, RegMixup performs \textbf{2.67\%} better than Mixup and \textbf{3.20\%} better than DNN. Similarly, on ImageNet-V2 it performs \textbf{1.08\%} better than Mixup and \textbf{1.13\%} more than DNN. RegMixup also outperforms AugMix on all the considered datasets, while is outperformed by DE on ImageNet-V2 (by \textbf{1.02\%}) and performs competitively on ImageNet-SK.  

\textit{These experiments clearly show the strong generalization of RegMixup under various in-distribution and CS experiments. They also show that it does not trade clean data accuracy to do so.}

\subsection{Out-of-Distribution Detection Experiments}
\label{sec:oodExperiments}

Following the standard procedure~\citep{Liu2020-sngp}, 
we report the performance in terms of AUROC\footnote{Area Under Receiver Operating Characteristic curve.} for the binary classification problem between in- and out-distribution samples. The predictive uncertainty of the model is typically used to obtain these curves. 
Given an uncertainty measure (normally entropy, refer Appendix \ref{sec:oodmetrics} for an extensive discussion), it is important for models to be more uncertain on OOD samples than on in-distribution samples to be able to distinguish them accurately. This behaviour would lead to a better AUROC.
\begin{table}
\centering
\resizebox{\linewidth}{!}
{
	\begin{tabular}{c|ccc|ccc|ccc|ccc}
		\toprule
		&
 		\multicolumn{6}{c}{\textbf{WRN28-10}} & \multicolumn{6}{c}{\textbf{RN50}} \\
		 	&
 		\multicolumn{3}{c}{\textbf{CIFAR10 (In-Distribution)}} & \multicolumn{3}{c}{\textbf{CIFAR100 (In-Distribution)}} &
 		\multicolumn{3}{c}{\textbf{CIFAR10 (In-Distribution)}} & \multicolumn{3}{c}{\textbf{CIFAR100 (In-Distribution)}} \\

		\cmidrule{2-13}
		
 		\textbf{Out-of-Distribution} &
 		\multicolumn{1}{c}{\textbf{C100}} & \multicolumn{1}{c}{\textbf{SVHN}} &
 		\multicolumn{1}{c}{\textbf{T-ImageNet}}
&
 		\multicolumn{1}{c}{\textbf{C10}} & \multicolumn{1}{c}{\textbf{SVHN}} &
 		\multicolumn{1}{c}{\textbf{T-ImageNet}} &
 		\multicolumn{1}{c}{\textbf{C100}} & \multicolumn{1}{c}{\textbf{SVHN}} &
 		\multicolumn{1}{c}{\textbf{T-ImageNet}}
&
 		\multicolumn{1}{c}{\textbf{C10}} & \multicolumn{1}{c}{\textbf{SVHN}} &
 		\multicolumn{1}{c}{\textbf{T-ImageNet}}
 		\\
\midrule
 		\textbf{Methods} & \multicolumn{3}{c}{\textbf{AUROC ($\uparrow$)}} & \multicolumn{3}{c}{\textbf{AUROC ($\uparrow$)}} & \multicolumn{3}{c}{\textbf{AUROC ($\uparrow$)}} & \multicolumn{3}{c}{\textbf{AUROC ($\uparrow$)}}    \\
 		\midrule
 DNN            & 88.61  & 96.00 & 86.44 &81.06 &  79.68 & 80.99 & 88.61 & 93.20 & 87.82                 & 79.33 & 82.45 & 79.89  \\
  Mixup          & 83.17 & 87.53 & 84.02          &78.37  & 78.68                        &  80.61 &84.24 & 89.40 & 84.89                & 77.02  &  76.86                  &  80.14     \\
 RegMixup (\textbf{our})     & 89.63 & \textbf{96.72} & \underline{\textbf{90.19 }}          &\textbf{81.27 }  &  \underline{\textbf{89.32 }}  & \textbf{83.13 }  &\textbf{89.63} &  \textbf{\underline{95.39} } & \textbf{90.04}       & \textbf{79.44 } & \textbf{\underline{88.66} } & \textbf{\underline{82.56}}      \\
 \cmidrule{2-13}

 DNN-SN         & 88.56 & 95.59  & 87.71    &81.10  &  83.43   & 82.26  & 88.19  & 93.46& 87.55               & 79.20  & 80.78  & 79.90 \\
 DNN-SRN        & 88.46  & 96.12  & 87.43        &81.26  &  85.51  & 82.41           & 88.82 & 93.54 & 87.82                & 78.77 & 82.39  & 79.70  \\
 SNGP           & \textbf{90.61 }  & 95.25  & 90.01                   &79.05   & 86.78           & 82.60   &-  &- &-                  &- &-  &-    \\
 KFAC-LLLA      & 89.33  & 94.17              & 87.81                   &81.04  & 80.32             &  81.57        &  89.54   & 93.13   &  88.32             & 79.30  &  82.80  &    80.17     \\
Aug-Mix & 89.78  & 91.3  & 88.99  & 81.10  & 76.64  & 80.56  &- &- &- &- &- &-\\
\midrule 
 DE (5$\times$) & \underline{91.25}  & \underline{97.53 } & 89.52             &\underline{83.26 } &  85.07   & \underline{83.40 }   & \underline{91.38}  & 96.90  &  90.5     & \underline{81.93 } & 85.08  & 82.15          \\

\midrule 
\end{tabular}%

}
\caption{Out-of-distribution detection results (\%) for WideResNet28-10 and ResNet50 for models trained on C10 and C100. See Appendix~\ref{sec:experimentDetails} for the cross-validated hyperparameters.}
\vspace{-2.5em}

	\label{tab:new_ood_wrn}
\end{table}

%

We report the OOD detection results for small-scale experiments (CIFAR 10/100) in Table~\ref{tab:new_ood_wrn}, and for large-scale experiments (ImageNet) in Table~\ref{tab:imagenet}. 
Compared to the single-model approaches, it is clear that RegMixup outperforms all the baselines in both small-scale and large-scale experiments, except only in one situation (C100 as OOD, trained on WRN) where SNGP outperformed it by an AUROC of $0.98$. In fact, on ImageNet experiments, RegMixup outperformed all the baselines including DE. We would also like to highlight that RegMixup provides significantly better performance compared to Mixup on all these experiments. For example, the improvement is more than $9\%$ when SVHN is treated as the OOD dataset for WRN trained on either C10 or C100 (refer Table~\ref{tab:new_ood_wrn}). Similarly, we can observe dramatic improvements on other experiments as well.

\section{Conclusive Remarks}
We proposed RegMixup, an extremely simple approach that combines Mixup with the standard cross-entropy loss. We conducted a wide range of experiments and showed that RegMixup significantly improved the reliability of uncertainty estimates of deep neural networks, while also provided a notable boost in the accuracy. We showed that RegMixup did not just outperform Mixup, it also outperformed most recent state-of-the-art approaches in providing reliable uncertainty estimates. 

We hope that our work opens possibilities to  explore situations where ERM and VRM can explicitly be combined together for practical benefits. An example would be label smoothing~\citep{labelsmoothing} that can be seen as an instance of VRM where the vicinal distribution is over the labels and the marginal distribution of the input samples (e.g., images) is approximated using deltas. In Appendix \ref{sec:additional} we conduct a preliminary analysis using CutMix \cite{CutMix} and recent techniques to train Vision Transformers \cite{ViT} that alternate between Mixup and CutMix. Another possible future direction would be to use bi-modal vicinal distribution along with importance sampling in order to ensure that the unperturbed samples as well are used during training with high probability. The observation that RegMixup ends up acting as a soft proxy to entropy maximizer for interpolated samples, a potential extension of our work regards the possibility of exploring different interpolation techniques to enforce high entropy behaviour on those regions of the input space. 

\section*{Acknowledgements}
This work is supported by the UKRI grant: Turing AI Fellowship EP/W002981/1 and EPSRC/MURI grant: EP/N019474/1. We would  like to thank the Royal Academy of Engineering, FiveAI and Meta AI. Francesco Pinto's PhD is funded by the European Space Agency (ESA). Meta AI authors are neither supported by the UKRI grants nor have any relationships whatsoever to the grant. Thanks to John Redford, Jon Sadeghi, Kemal Oksuz, Zygmunt Lenyk, Guillermo Ortiz-Jimenez, Amartya Sanyal, Pau de Jorge, and David Lopez-Paz for their valuable comments on the paper.

\newpage

\bibliographystyle{plainnat}
\bibliography{example_paper.bib}

\section*{Checklist}



\begin{enumerate}

\item For all authors...
\begin{enumerate}
  \item Do the main claims made in the abstract and introduction accurately reflect the paper's contributions and scope?
    \answerYes{}
  \item Did you describe the limitations of your work?
    \answerYes{See the "Methods considered for Comparison" section}
  \item Did you discuss any potential negative societal impacts of your work?
    \answerNo{Not expected}
  \item Have you read the ethics review guidelines and ensured that your paper conforms to them?
    \answerYes{}
\end{enumerate}

\item If you are including theoretical results...
\begin{enumerate}
  \item Did you state the full set of assumptions of all theoretical results?
    \answerNA{}
        \item Did you include complete proofs of all theoretical results?
    \answerNA{}
\end{enumerate}

\item If you ran experiments...
\begin{enumerate}
  \item Did you include the code, data, and instructions needed to reproduce the main experimental results (either in the supplemental material or as a URL)?
    \answerYes{Code is on github. The appendix contains an explicative pseudocode}
  \item Did you specify all the training details (e.g., data splits, hyperparameters, how they were chosen)?
    \answerYes{They are specified either in the Experiments section of the main paper or the Appendix}
        \item Did you report error bars (e.g., with respect to the random seed after running experiments multiple times)?
    \answerNA{}
        \item Did you include the total amount of compute and the type of resources used (e.g., type of GPUs, internal cluster, or cloud provider)?
    \answerYes{See Appendix}
\end{enumerate}

\item If you are using existing assets (e.g., code, data, models) or curating/releasing new assets...
\begin{enumerate}
  \item If your work uses existing assets, did you cite the creators?
    \answerYes{}
  \item Did you mention the license of the assets?
    \answerNo{}
  \item Did you include any new assets either in the supplemental material or as a URL?
    \answerNo{}
  \item Did you discuss whether and how consent was obtained from people whose data you're using/curating?
    \answerNA{}
  \item Did you discuss whether the data you are using/curating contains personally identifiable information or offensive content?
    \answerNA{}
\end{enumerate}

\item If you used crowdsourcing or conducted research with human subjects...
\begin{enumerate}
  \item Did you include the full text of instructions given to participants and screenshots, if applicable?
    \answerNA{}
  \item Did you describe any potential participant risks, with links to Institutional Review Board (IRB) approvals, if applicable?
    \answerNA{}
  \item Did you include the estimated hourly wage paid to participants and the total amount spent on participant compensation?
    \answerNA{}
\end{enumerate}

\end{enumerate}

\newpage
\appendix
\onecolumn

\newpage

\appendix

\section{Calibration on In-distribution and Covariate-shift datasets}
\label{sec:miscalibr}

\begin{wraptable}{r}{0.5\textwidth}

\resizebox{\linewidth}{!}
{
	\begin{tabular}{ccc|cc}
		\toprule
		& \multicolumn{4}{c}{\textbf{IND }} \\

& \multicolumn{2}{c|}{\textbf{WRN28-10}} & \multicolumn{2}{c}{\textbf{RN50}} \\
		& \multicolumn{1}{c}{\textbf{C10 (Test)}} & \multicolumn{1}{c|}{\textbf{C100 (Test)}} & \multicolumn{1}{c}{\textbf{C10 (Test)}} & \multicolumn{1}{c}{\textbf{C100 (Test)}}  \\

		\textbf{Methods} & \multicolumn{2}{c|}{\textbf{AdaECE ($\downarrow$)}} & \multicolumn{2}{c}{\textbf{AdaECE ($\downarrow$)}} \\
		\midrule

DNN               & 1.34  & 3.84 & 1.45   & 2.94 \\
Mixup           & 1.16   & 1.98 & 2.17    & 7.47     \\
RegMixup (\textbf{our})  & \textbf{\underline{0.50} }   & \textbf{\underline{1.76} } & \textbf{\underline{0.94}}  & \textbf{\underline{1.53}} \\
\cmidrule(r){2-5}                                                                                                

SNGP            & 0.87   & 1.94 & - & - \\
Augmix & 1.67  & 5.54 & - & - \\ 
\midrule
DE (5$\times$)  & 1.04  & 3.29 & 1.28  & 2.98 \\
\bottomrule 
\end{tabular}%

}
\caption{CIFAR IND calibration performance (\%).}
	\label{tab:ece_indc10c100}
\resizebox{\linewidth}{!}{

\begin{tabular}{cc|cccc}
	\toprule
	 & \multicolumn{1}{c|}{\textbf{IND}}  & \multicolumn{4}{c}{\textbf{Covariate Shift}}  \\

	& \multicolumn{1}{c|}{\textbf{ImageNet-1K (Test)}} & \multicolumn{1}{c|}{\textbf{ImageNet-R}} &
	\multicolumn{1}{c|}{\textbf{ImageNet-A}} &
	\multicolumn{1}{c|}{\textbf{ImageNet-V2}} &
	 \multicolumn{1}{c}{\textbf{ImageNet-Sk}}   \\

	  & \textbf{AdaECE} ($\downarrow$)  & \textbf{AdaECE} ($\downarrow$) & \textbf{AdaECE} ($\downarrow$) & \textbf{AdaECE} ($\downarrow$) & \textbf{AdaECE} ($\downarrow$)    \\

\midrule 
DNN           & 1.81 & 13.56 & 44.90 & 4.13 & 14.48 \\ 
Mixup     & \underline{\textbf{1.29}} & 12.08 & 44.63 & 4.28 & 15.26 \\
RegMixup (\textbf{our})  & 1.37 & 13.30 & \underline{\textbf{41.18}} & \underline{\textbf{3.38}} & 15.35 \\ 
\cmidrule(r){2-6}                                                                                                

AugMix    & 2.05 & \underline{\textbf{11.24}} & 42.83 & 3.94 & \underline{\textbf{14.26}} \\ 
\midrule
DE (5$\times$) & 1.38 & 13.55 & 42.88 & 4.02 & 17.32 \\ 

\bottomrule
	\end{tabular}%
}
    \caption{ImageNet calibration performance on IND and CS datasets (\%).}
    \label{tab:imagenet_ece}
\resizebox{\linewidth}{!}
{
	\begin{tabular}{ccccc|cccc}
		\toprule
		& \multicolumn{8}{c}{\textbf{Covariate Shift}} \\
& \multicolumn{4}{c|}{\textbf{WRN28-10}} &  \multicolumn{4}{c}{\textbf{R50}} \\ 
		& \multicolumn{1}{c}{\textbf{C10-C}} &  \multicolumn{1}{c}{\textbf{C10.1}} &  \multicolumn{1}{c}{\textbf{C10.2}} &  \multicolumn{1}{c|}{\textbf{C100-C}} & \multicolumn{1}{c}{\textbf{C10-C}} &  \multicolumn{1}{c}{\textbf{C10.1}} &  \multicolumn{1}{c}{\textbf{C10.2}} &  \multicolumn{1}{c}{\textbf{C100-C}} \\

		\textbf{Methods} & \multicolumn{4}{c|}{\textbf{AdaECE ($\downarrow$)}} & \multicolumn{4}{c}{\textbf{AdaECE ($\downarrow$)}}  \\
		\midrule

DNN              & 12.62  & 4.13  & 8.81  & 9.94   & 12.29  & 4.36 & 8.89 & 19.76 \\
Mixup           & 7.93 & 4.39   & 7.44 & 10.45 & \textbf{\underline{10.75}} & 5.72 & 10.59 & 12.63  \\
RegMixup (\textbf{our})     & 9.08  & \textbf{\underline{2.57} } & \textbf{\underline{6.83} } & \textbf{\underline{7.93}}  & 11.37 & \textbf{\underline{2.89}} & \textbf{\underline{6.74}} & \textbf{\underline{11.47}}      \\
\cmidrule{2-9}
SNGP             & 11.34   & 4.36   & 8.33 & 10.43 & - & - & - & -    \\
AugMix & \textbf{\underline{4.56}}  & 3.23  & 8.33  & 12.15 & - & - & - & - \\
\midrule 
DE (5 $\times$)   & 10.31  & 2.60  & 7.50  & 12.36   & 12.68 & 4.10 & 6.94 & 12.36      \\
\bottomrule 
\end{tabular}%

}
\caption{CIFAR CS calibration performance (\%).}
	\label{tab:ece_dshift}
\end{wraptable}
Additionally, we provide the calibration performance of various competitive approaches. Briefly, calibration quantifies how similar a model's confidence and its accuracy are~\citep{CalibrationReview}). To measure it, we employ
the recently proposed Adaptive ECE (AdaECE)~\citep{mukhoti2020calibrating}.
For all the methods, the AdaECE is computed after performing temperature scaling~\citep{GuoCalibration} with a cross-validated temperature parameter. We also provide the AdaECE without temperature scaling in Appendix~\ref{sec:additionalExperiments}.  For completeness, we also report the ECE in Appendix \ref{sec:ece}.

In terms of calibration on in-domain test sets (refer Tables~\ref{tab:ece_indc10c100} and ~\ref{tab:imagenet_ece}), our method either remarkably improves the AdaECE with respect to Mixup and DNN, or performs competitively (on ImageNet-1K).

Under covariate shift (refer Tables~\ref{tab:imagenet_ece} and ~\ref{tab:ece_dshift}), on corrupted inputs, RegMixup underperforms with respect to Mixup on C10-C, but not on C100-C. On all other C10 covariate shift datasets, RegMixup outperforms both Mixup and DNN.
Considering also the other baselines, except for the case of C10-C (in which AugMix significantly outperforms any other baseline on WRN28-10), our method provides the best calibration in all other cases.
 For example, on C100-C experiments on WRN28-10, in terms of AdaECE, RegMixup obtains a \textbf{4.43\%} improvement over DE, \textbf{2.52\%} over Mixup, and \textbf{2.47\%} over SNGP. Though RegMixup outperformed all other approaches in 12 scenarios out of total 17 presented here, it is clear that there is no single method that outperforms any other in all the considered settings.

\section{Experimental Details}
\label{sec:experimentDetails}

\subsection{Code-base}
The \textbf{RegMixup} training procedure is outlined in Algorithm~\ref{alg:RegMixup}.

For fair comparisons, when training on C10 and C100, we developed our own code base for all the approaches (except SNGP, DUQ and AugMix) and performed an extensive hyperparameter search to obtain the strongest possible baselines. 

We would like to highlight that it was not easy to make a few recent state-of-the-art approaches work in situations different from the ones they reported in their papers as these approaches mostly required  non-trivial changes to the architectures and additional sensitive hyperparametes. We also observed that their performances did not easily translate to new situations. Below we highlight few of these issues we faced and the measures we took for comparisons.

\textbf{For DUQ}, the original paper did not perform large scale experiments similar to ours. Unfortunately, we could not manage to make their code work on C100 as the training procedure seemed to be \textit{unstable}. For this reason, \textit{wherever possible}, we borrowed the numbers for DUQ from the SNGP paper. Please note that the authors of SNGP performed non-trivial modifications to the original DUQ methodology to make it work on C100.

\textbf{For SNGP}, we used the publicly available code following exactly the same procedure as mentioned in their original paper. The code \textit{diverges slightly} from the procedure described in their paper, hence the slight differences in the performance. The only modification we performed to the official code-base was to make the inference procedure consistent with the one described in the paper: indeed, in their code they implement a mean-field approximation to estimate the predictive distribution ~\citep{lu2020Jackknife}, while in their paper they use Monte Carlo Integration with a number of samples equal to the number of members in the ensembles they use as a baseline, which provides better calibration. The rationale is that we could not find an obvious way to tune the mean-field approximation hyperparameters to improve at the same time both the calibration and OOD detection performance (indeed, \textit{the mean-field approximation imposes a trade-off between calibration and OOD detection performance}). Additionally, since the standard KFAC-LLLA uses the same Monte Carlo Integration procedure, we opted for the latter for a fair comparison. For the SNGP RN50 experiments, we tried running the official implementation on C10 and C100, but we could not make SNGP converge to SOTA accuracy values. The authors of SNGP did not provide experiment results on C10 and C100 on RN50. Hence we decided \textit{not to report} these experiments for SNGP.

\textbf{For the KFAC-LLLA} we leverage the official repository\footnote{\url{https://github.com/19219181113/LB_for_BNNs}}~\citep{laplacebridge} and the Backpack library ~\citep{backpack} for the computation of the Kronecker-Factored Hessian.

\textbf{For AugMix}, we used their code base and the exact training procedure. AugMix seems to be sensitive to hyperparameters of the training procedure as we could could not get the considered architectures to converge to acceptable accuracy levels under the training regime we used for \textbf{all} other baselines. Even with the recipes provided in the AugMix paper, we could not get it to converge to competitive accuracy levels when using RN50 on C10 and C100 hence we decided \textit{not to report} these experiments for AugMix.

\begin{algorithm}[tb]
  \caption{RegMixup training procedure}
 \label{alg:RegMixup}
\begin{algorithmic}
\INPUT{Batch $\mathcal{B}$, $\alpha$, $\theta_t$}
  \STATE $\bar{\mathcal{B}} \gets \emptyset$\
  \STATE  $\lambda_0 \sim \texttt{Beta}(\alpha,\alpha)$\;
  \FOR{$\forall (\bfx_i, \bfy_i) \in \mathcal{B}$}
  \STATE  randomly select $(\bfx_j,\bfy_j)$ from $\mathcal{B} \backslash (\bfx_i, \bfy_i)$
  \STATE  $\bar{\mathcal{B}} \gets \bar{\mathcal{B}} \cup (\lambda_0 \bfx_i +(1-\lambda_0) \bfx_j, \lambda_0 \bfy_i + (1-\lambda_0)\bfy_j)$\;
  \ENDFOR
     \STATE $\mathcal{L} = \textsc{CE} (\mathcal{B}) + \textsc{CE} (\bar{\mathcal{B}})$\; \textit{// Loosely speaking, compute cross-entropy loss on both the batches}
      \STATE  return $\theta_{t+1}$ obtained by updating $\theta_t$ by optimizing the above loss
\end{algorithmic}
\end{algorithm}
\subsection{Optimization}
For\textbf{ C10 and C100} training, we use SGD with Nesterov momentum $0.9$ for 350 epochs and a weight decay of $5 \times 10^{-4}$. For WRN, we apply a dropout $p=0.1$ at train time. 
For all our experiments we set the batch size to 128\footnote{For SNGP and DUQ, we use the hyperparameters suggested in their original papers.}. At training time, we apply standard augmentations \texttt{random crop} and \texttt{horizontal flip} similar to~\citep{Liu2020-sngp}). The data is appropriately normalized before being fed to the network both at train and test time.  

For \textbf{ImageNet-1K} training, we use SGD with momentum for 100 epochs, learning rate $0.1$, cosine learning scheduler, weight decay of $1 \times 10^{-4}$, batch size 128 and image size $224 \times 224$. We use \texttt{color jitter}, \texttt{random horizontal flip} and \texttt{random crop} for augmentation. We leverage the \texttt{timm} library for training \citep{timm} all the considered methods with Automatic Mixed Precision to accelerate the training.

\subsection{Hyperparameters}
\label{sec:cv}

\begin{compactitem}
    \item For DNN-SN and DNN-SRN the spectral norm clamping factor (maximum spectral norm of each linear mapping)  $c \in \{0.5, 0.75, 1.0\}$ and the target of stable rank $r \in \{0.3,0.5,0.7,0.9\}$ (as $r=1$ for SRN is the same as applying SN with $c=1.0$). Refer to~\citet{miyato2018spectral} and~\citep{Sanyal2020-SRN} for details about these hyperparameters.
    \item For Mixup, we consider a wide range of Beta distribution hyperparameter $\alpha \in \{0.1,0.2,0.3,0.4,0.5, 1, 5, 10, 20\}$.
    \item For RegMixup we consider the Beta distribution hyperparameters to be $\alpha \in \{0.1, 0.2, 0.3, 0.4, 0.5, 1, 5, 10, 15, 20, 30\}$, and the mixing weight $\eta \in \{0.1, 1, 2\}$. 
    \item For KFAC-LLLA we take $1000$ samples from the distribution. Although the number might seem quite high, we could not notice significant improvements using a lower number of samples. We tuned the prior variance $\sigma_0$ needed for the computation of the Laplace approximation minimising the ECE on the validation set. We also tried using the theoretical value $\sigma_0=1/\tau$~\citep{HeinBeingBayesianABit2020}, where $\tau$ represents the weight decay, but it produced inferior results with respect to our cross-validation procedure. We provide an overview of the KFAC-LLLA in Appendix~\ref{sec:BayesianTest}.
    \item For Deep Ensembles we use 5 members.
    \item When temperature scaling is applied, the temperature $T$ is tuned on the validation set, minimising the ECE (we considered values ranging from $0.1$ to $10$, with a step size of $0.001$). For Deep Ensembles, we first compute the mean of the logits, then scale it by the temperature parameter before passing it through the softmax.
\end{compactitem}

All the cross-validated hyperparameters are reported in Table \ref{tab:hyperparams}. The cross-validation is performed with stratified-sampling on a 90/10 split of the training set to maximise accuracy\footnote{Except for the $\sigma_0$ of the KFAC-LLLA, as we could not observe significant differences in Accuracy between hyperparameters optimising the accuracy and ECE} on C10 and C100. For ImageNet, we split the test set using the same proportion to obtain the validation set, which is then removed from the test set during evaluation. It is important to observe that:
\begin{compactitem}
    \item Cross-validating hyperparameters based solely on the ECE can prefer models with lower accuracy but better calibration. However, a method improving calibration should avoid degrading accuracy.  
    \item Hyperparameters should not be cross-validated based on CS experiments and OOD detection metrics as they these datasets should be unknown during the training and hyperparameter selection procedure as well.
\end{compactitem}

The results of the cross-validation for RegMixup for CIFAR-10, CIFAR-100 can be found in Table  \ref{tab:cv_regmixup}. To support our claim that Mixup tends to prefer lower $\alpha$ values, we do not only report the cross-validation accuracy on Mixup for the experiments in the main paper (Table \ref{tab:bad_mixup}) but we also report them for two other popular architectures on CIFAR-10 and CIFAR-100: DenseNet-121 \cite{DenseNet} and PyramidNet200 \cite{PyramidNet} (Table \ref{tab:pyr_dens}). 
\begin{table}[]
    \centering
    \begin{tabular}{c|c|ccccc}
        Train Data/ & Hyperparas & \multicolumn{2}{c}{\textbf{C10}} & \multicolumn{2}{c}{\textbf{C100}} & \multicolumn{1}{c}{\textbf{ImageNet}} \\
        Architecture & & WRN & R50 & WRN & R50 \\
        \midrule 
        DNN & $T$ & 1.32 & 1.51 & 1.33 & 1.42 & 1.19 \\ 
                \hline
        \multirow{2}{*}{DNN-SN}
        & $c$ & 0.5 & 0.5 & 0.5 & 0.5 &-\\
        & $T$ & 1.42 & 1.51 & 1.21 & 1.42 &-\\
        \hline
        \multirow{2}{*}{DNN-SR}
        & $r$ & 0.3 & 0.3 & 0.3 & 0.3  &-\\ 
        & $T$ & 1.33 & 1.41 & 1.22 & 1.42  &-\\
        \hline
        DE & $T$ & 1.31 & 1.42  & 1.11 & 1.21 & 1.29\\ 
                \hline

        SNGP & $T$ & 1.41 & - & 1.52 & -  & -\\ 
        \hline

        \multirow{2}{*}{Mixup}
        & $\alpha$ & 0.3 &  0.3 & 0.3 & 0.3 & 0.1 \\ 
        & $T$ & 0.73 & 0.82 & 1.09 & 1.21 & 1.06\\
        \hline
        \multirow{2}{*}{RegMixup}
        & $\eta = 1$, $\alpha$ & 20 &  20 & 10 & 10 & 10 \\ 
        & $T$ & 1.12 & 1.31 & 1.23 & 1.21 & 1.14 \\
        \hline
        \multirow{2}{*}{KFAC-LLLA}
        & \#samples & 1000 & 1000 & 1000 & 1000\\ 
        & $\sigma_0$ & 1 & 0.6 & 4 & 0.1 &-\\ 
        \hline

    \end{tabular}
    \caption{Cross-validated hyperparameters. Note, $T$ and $\sigma_0$ are cross-validated by minimizing the ECE. All other hyperparameters have been tuned to maximise the accuracy.}
    \label{tab:hyperparams}
\end{table}

\section{Existing Uncertainty Measures}
\label{sec:oodmetrics}
There are various uncertainty measures and there is no clear understanding on which one would be more reliable. In our experiments we considered the following metrics and chose the one best suited for each method in order to create the strongest possible baselines. Let $K$ denote the number of classes, $\bfp_i$ the probability of $i$-th class, and $\bfs_i$ the \texttt{logit} of $i$-th class. Then, these uncertainty measures can be defined as:
\begin{compactitem}
    \item {\bf Entropy}: $H(\bfp(\bfx)) = - \sum_{i=1}^{K} \bfp_i \log \bfp_i$.
    \item {\bf Dempster-Shafer}~\citep{evidentialDempsterShafer}: $\texttt{DS}(\bfx) = \nicefrac{K}{(K + \sum_{i=1}^{K} \exp(\bfs_i))}$.
    \item {\bf Energy}: $E(\bfx) = - \log \sum_{i=1}^{K} \exp(\bfs_i)$ (ignoring the temperature parameter). This metric was used in~\citep{liu2020energybased} for OOD.
    \item {\bf Maximum Probability Score}: $\texttt{MPS}(\bfx) = \max_i \bfp_i$.
    \item {\bf Feature Space Density Estimation} (FSDE): Assuming that the features of each class follow a Gaussian distribution, there are several ways one can estimate the {\em belief} of a test sample belonging to in-distribution data and treat it as a measure of uncertainty. One such approach is to compute the Mahalanobis score $\arg \min_{i \in y} (\phi(\bfx)- {\mu}_i)^T \Sigma_i^{-1}(\phi(\bfx)- \mu_i)$, where $\mu_i$ and $\Sigma_i$ are class-wise mean and the covariance matrices of the {\em train} data, and $\phi(\bfx)$ is the feature vector.
\end{compactitem}

\textit{In the main paper, we report the OOD detection performance using the \texttt{DS} score (as it provided slightly improved performance in most cases), except when it damages the performance of a method (e.g. Mixup) or when it does not yield improvements (e.g. KFAC-LLLA). In these situations we use the entropy as the uncertainty measure.}

\paragraph{Remarks regarding various metrics:} We would like to highlight a few important observations that we made regarding these metrics. \textbf{(1)} \textbf{$\texttt{DS}$ and $E$ are equivalent} as they are both decreasing functions of $\sum_{i=1}^{K} \exp(\bfs_i)$, and since $\log$ does not modify the monotonicity, both will provide the same ordering of a set of samples. Hence, will give the same AUROC values. \textbf{(2)} We observed $\texttt{DS}$ and $H$ to perform similarly to each other except in a few situations where $\texttt{DS}$ provided slightly better results. \textbf{(3)} $\texttt{MPS}$, in many situations, was slightly worse. \textbf{(4)} We found Gaussian assumption based density estimation to be \textbf{unreliable}. Though it provided extremely competitive results for C10 experiments, sometimes slightly better than the $\texttt{DS}$ based scores, it performed very poorly on C100. We found this score to be highly unstable as it involves large matrix inversions. We applied the well-known tricks such as perturbing the diagonal elements and the low-rank approximation with high variance-ratio, but the results were sensitive to such stabilization and there is no clear way to cross-validate these hyperparameters. 

\section{Calibration Metrics without Temperature Scaling}
\label{sec:additionalExperiments}

For completeness, we report the calibration metrics over all the methods and considered datasets without the temperature scaling~\citep{GuoCalibration} in Tables~\ref{tab:r50c10_notscaling} and~\ref{tab:imagenet_ece_notemp}. 
Details about the cross-validation procedure used when temperature scaling is applied is provided in Appendix~\ref{sec:experimentDetails}.
\begin{table*}
\centering
\scalebox{0.7}{
\begin{small}

\begin{tabular}{cc|c|c|c}
		\toprule
		\textbf{Methods} & \multicolumn{1}{c}{\textbf{Clean}} & \multicolumn{1}{c}{\textbf{CIFAR-10-C}} & \multicolumn{1}{c}{\textbf{CIFAR 10.1}} & \multicolumn{1}{c}{\textbf{CIFAR 10.2}}\\

		 & \textbf{AdaECE} ($\downarrow$)
		& \textbf{AdaECE} ($\downarrow$)
		& \textbf{AdaECE} ($\downarrow$)
        & \textbf{AdaECE} ($\downarrow$)

		 \\
		\midrule
\textbf{C10 R50} \\
		\midrule
DNN               & 3.02  & 17.30  & 7.39  & 12.24  \\
Mixup           & 2.87 & \textbf{\underline{11.35}} & \textbf{\underline{4.05}} & \textbf{\underline{7.72}} \\
RegMixup (\bf{Ours})  & 1.40 & 11.51 & 4.15 & 8.23 \\

\midrule
DE (5$\times$)   & 2.10 & 13.99 & 6.23 & 10.33  \\

\midrule
\\ 
\\ 
		\midrule

\textbf{C10 WRN} \\
		\midrule

DNN              & 2.27 & 15.92 & 6.00 & 11.00  \\
Mixup           & 2.23 & \textbf{\underline{7.93}} & 7.22  & \textbf{\underline{6.58}} \\
RegMixup (\bf{Ours})   & \textbf{\underline{ 0.67}} & 8.36 & \textbf{\underline{3.02}} & 7.03 \\
\midrule

SNGP         & 1.51  & 11.33 & 5.59 & 10.85 \\

AugMix & 1.89 & \textbf{\underline{5.77}} & 4.10 & 9.61  \\ 

\midrule
DE (5$\times$)  & 1.74 & 13.52 & 4.33 & 9.44 \\

\midrule

\end{tabular}
\;
\begin{tabular}{cc|c}
	\toprule
	\textbf{Methods} & \multicolumn{1}{c}{\textbf{Clean}} & \multicolumn{1}{c}{\textbf{CIFAR-100-C}} \\

	 & \textbf{AdaECE} ($\downarrow$)
	 & \textbf{AdaECE} ($\downarrow$) \\
	 \midrule 
	 \textbf{C100 R50} \\
	\midrule

DNN               &  9.47   & 25.17    \\
Mixup            & 7.47   & 21.52 \\
RegMixup (\bf{Ours})  & 3.92  & 13.68  \\
\midrule
DE (5$\times$)    & 6.50 & 19.76  \\

\midrule
\\ 
\\ 

		\midrule
		        \textbf{C100 WRN} \\
		\midrule

DNN              & 5.30  & 17.38   \\
Mixup            & 3.60  & 16.54    \\
RegMixup (\bf{Ours})      & 2.47  & 10.49         \\
\midrule

SNGP         & 5.65 & 10.89  \\

AugMix & 5.23 & 13.67 \\ 

\midrule
DE (5$\times$)   & 3.92 & 13.47     \\

\midrule
    \end{tabular}

\end{small}
}
\caption{CIFAR calibration performance (\%) without temperature scaling}
	\label{tab:r50c10_notscaling}
\end{table*}

\begin{table}

\resizebox{\linewidth}{!}{

\begin{tabular}{cc|cccc}
	\toprule
	 & \multicolumn{1}{c|}{\textbf{IND}}  & \multicolumn{4}{c}{\textbf{Covariate Shift}}  \\

	& \multicolumn{1}{c|}{\textbf{ImageNet-1K (Test)}} & \multicolumn{1}{c|}{\textbf{ImageNet-R}} &
	\multicolumn{1}{c|}{\textbf{ImageNet-A}} &
	\multicolumn{1}{c|}{\textbf{ImageNet-V2}} &
	 \multicolumn{1}{c}{\textbf{ImageNet-Sk}}   \\

	  & \textbf{AdaECE} ($\downarrow$)  & \textbf{AdaECE} ($\downarrow$) & \textbf{AdaECE} ($\downarrow$) & \textbf{AdaECE} ($\downarrow$) & \textbf{AdaECE} ($\downarrow$)    \\

\midrule 
DNN           & 4.90 & 20.48 & 52.30 & 9.58 & 22.94 \\ 
Mixup     & \underline{\textbf{2.28}} & \underline{\textbf{14.70}} & 47.41 & 6.46 & \underline{\textbf{18.26}} \\
RegMixup (\textbf{our})  & 3.06 & 17.42 & \underline{\textbf{45.65}} & 7.34 & 20.85 \\ 
\cmidrule(r){2-6}                                                                                                

AugMix    & 4.28 & 19.13 & 51.35 & \underline{\textbf{3.94}} & 21.25 \\ 
\midrule
DE (5$\times$) & 3.61 & 17.32 & 51.64 & 7.94 & 19.35 \\ 

\bottomrule
	\end{tabular}%
}
    \caption{ImageNet calibration performance (\%) without temperature scaling.}
    \label{tab:imagenet_ece_notemp}
\end{table}

\section{Bayesian at Test Time: Last Layer Laplace Approximation}
\label{sec:BayesianTest}
A structural problem of using MLE logistic regression is that the produced uncertainties depend on the decision boundary. On the other hand, replacing the MLE logistic regression with a Bayesian logistic regression and estimating the predictive posterior employing a Laplace approximation allows to produce better uncertainties~\citep{HeinBeingBayesianABit2020}. However, a Bayesian training either requires a modification in the architecture~\citep{Liu2020-sngp} or makes the inference procedure very expensive~\citep{BayesDropout1,BayesDropout2}. Since the objective is to utilize the standard MLE training of neural networks, the idea of Kronecker-Factored Last Layer Laplace Approximation~\citep{HeinBeingBayesianABit2020} is making the network \textbf{Bayesian at test time} with almost no additional cost.  

Let $\bfw$ be the parameters of the of the last layer of a neural network, then we seek to obtain the posterior only over $\bfw$. Let $p(\bfw|\bfx)$ be the posterior, then the predictive distribution can be written as:
\begin{equation}
    \label{eq:predictive_int}
    p(y=k|\bfx, \dataset) = \int \texttt{softmax}(\bfs_k) p(\bfw|\dataset)d\bfw,
\end{equation}
where, $\bfs$ is the logit vector and $\texttt{softmax}(\bfs_k)$ is the $k$-th index of the $\texttt{softmax}$ output of the network.

The Laplace approximation assumes that the posterior $p(\bfs|\dataset) \sim \mathcal{N}(\bfs|\mu, \Sigma)$, where $\mu$ is a mode of the posterior $p(\bfw|\dataset)$ (found via standard optimization algorithms for NNs) and $\Sigma$ is the inverse of the Hessian $\hmat^{-1} = - (\nabla^2 \log p(\bfw| \dataset)|_{\mu})^{-1}$. For the formulations and definitions, including the variants with the terms associated to the bias, we refer to~\citep{HeinBeingBayesianABit2020}.

For our experiments, we obtain $\Sigma$ using the Kronecker-factored (KF) approximation~\citep{Ritter2018aLaplace}.
Broadly speaking, the KF approximation allows to reduce the computational complexity of computing the Hessian by factorizing the inverse of the Hessian as $\hmat^{-1} \approx \vmat^{-1} \otimes \umat^{-1}$, then the  covariance of the posterior evaluated at a point $\bfx$ takes following form $\Sigma = (\phi(\bfx)^T \vmat \phi(\bfx)) \umat$. This procedure can be easily implemented using the Backpack library~\citep{backpack} to compute $\vmat$ and $\umat$ by performing a single pass over the training set after the end of the training, as detailed in the Appendix of~\citep{HeinBeingBayesianABit2020} and clearly exemplified in the code-base of~\citep{laplacebridge}.

Let $\Sigma_k$ be the covariance matrix of the posterior over the last linear layer parameters for the k-th class obtained using the Laplace approximation around $\mu$, then, given an input $\bfx$, we obtain $\sigma_k = \phi(\bfx)^\top \Sigma_k \phi(\bfx)$ representing the variance of k-th logit $\bfs_k$. Once we obtain the covariance matrix, the Monte Carlo approximation of the predictive distribution (equation~(\ref{eq:predictive_int})) is obtained as:
\begin{align}
\label{eq:mcSoftmax}
    \tilde{p} = \frac{1}{m} \sum_{i=1}^m \texttt{softmax}(\bfs(i)),
\end{align}
where, $m$ logit vectors $\bfs(i)$ are sampled from a distribution with mean $\bfs$ and a covariance matrix (depending on the approximation used). Lu et. al~\citep{lu2020Jackknife} showed that similar performance can be achieved via the mean-field approximation which provides an approximate closed form solution of the integration in equation~(\ref{eq:predictive_int}) involving the re-scaling of the logits and then taking the softmax of the re-scaled logit. The re-scaling is defined as follows:
\begin{align}
\label{eq:mf0Logit}
    \tilde{\bfs}_k = \frac{\bfs_k}{\sqrt{1 + \lambda \sigma_k^2}}
\end{align}
Note, the scaling of the k-th logit depends on its variance (obtained using the Laplace approximation) and a hyperparameter $\lambda$. This approximation is efficient in the sense that it does not require multiple samples as required in the MC approximation (which can become expensive as the number of classes and samples grow).
In our experiments, we use the MC approximation, since we could not find an obvious way to fine-tune $\lambda$. Additionally, we observe that the mean-field approximation imposes a trade-off between calibration and OOD detection performance. Increasing $\lambda$, indeed, flattens the softmax distribution and improves OOD detection scores; although, as a consequence, harms calibration by making the network underconfident. 

\section{Additional Insights: RegMixup encourages compact and separated clusters}
\label{sec:fisher}
Here we provide additional experiments to show that RegMixup encourages more compact and separated clusters in the feature space.
We use the well known Fisher criterion~\citep[Chapter~4]{bishopPRML} to quantify the compactness and separatedness of the feature clusters.

\textbf{Fisher Criterion:} Let $\mathcal{C}_k$ denotes the indices of samples for $k$-th class. Then, the overall {\em within-class} covariance matrix is computed as $\mathbf{S}_W = \sum_{k=1}^K \mathbf{S}_k$, where $\mathbf{S}_k = \sum_{n \in \mathcal{C}_k} (\phi(\bfx_n) - \mu_k)(\phi(\bfx_n) - \mu_k)^\top$, $\mu_k =  \sum_{n \in \mathcal{C}_k} \frac{\phi(\bfx_n)}{N_k}$, and $\phi(\bfx_n)$ denote the feature vector. Similarly, the {\em between-class} covariance matrix can be computed as $\mathbf{S}_B = \sum_{k=1}^K N_k(\mu_k - \mu)(\mu_k - \mu)^\top$, where $\mu = \frac{1}{N} \sum_{k=1}^K N_k \mu_k$, and $N_k$ is the number of samples in $k$-th class. Then, the Fisher criterion is defined as $\alpha = \texttt{trace}(\mathbf{S}_W^{-1}\mathbf{S}_B)$. 

\begin{figure}
     \centering
     \begin{subfigure}[b]{0.3\textwidth}
         \centering
         \includegraphics[height=0.15\textheight]{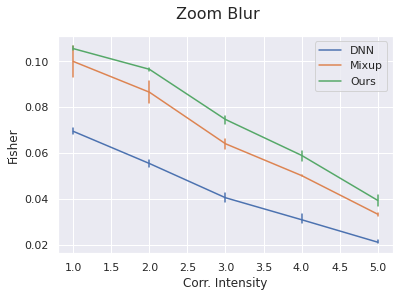}
     \end{subfigure}
     \hfill
     \begin{subfigure}[b]{0.3\textwidth}
         \centering
         \includegraphics[height=0.15\textheight]{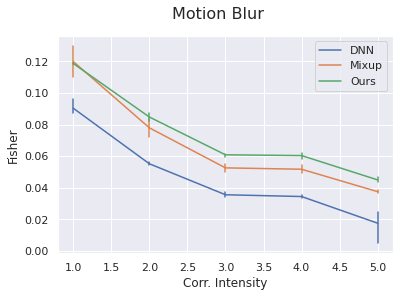}
     \end{subfigure}
     \hfill
     \begin{subfigure}[b]{0.3\textwidth}
         \centering
         \includegraphics[height=0.15\textheight]{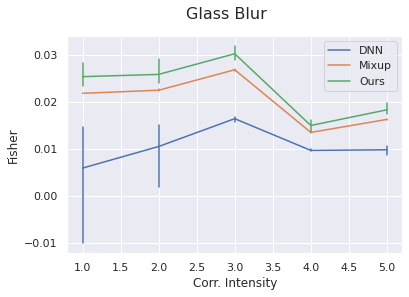}
     \end{subfigure} \\
     \begin{subfigure}[b]{0.3\textwidth}
         \centering
         \includegraphics[height=0.15\textheight]{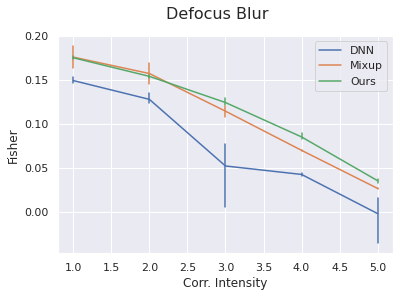}
     \end{subfigure}
     \hfill
     \begin{subfigure}[b]{0.3\textwidth}
         \centering
         \includegraphics[height=0.15\textheight]{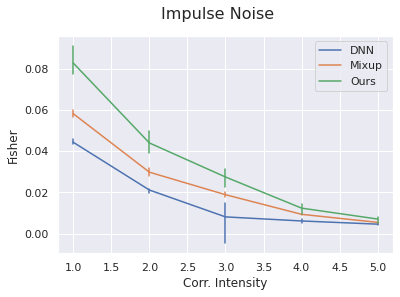}
     \end{subfigure}
     \hfill
     \begin{subfigure}[b]{0.3\textwidth}
         \centering
         \includegraphics[height=0.15\textheight]{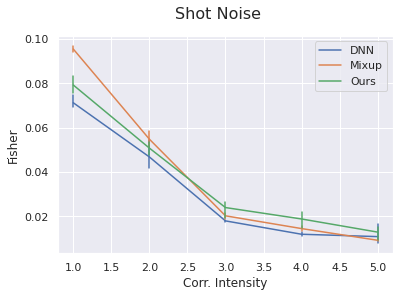}
     \end{subfigure} \\
     \begin{subfigure}[b]{0.3\textwidth}
         \centering
         \includegraphics[height=0.15\textheight]{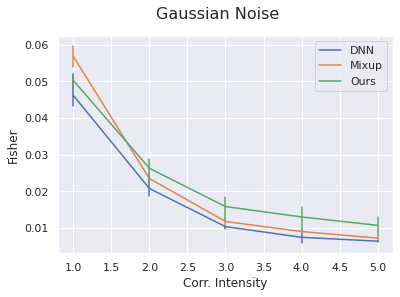}
     \end{subfigure}
     \hfill
     \begin{subfigure}[b]{0.3\textwidth}
         \centering
         \includegraphics[height=0.15\textheight]{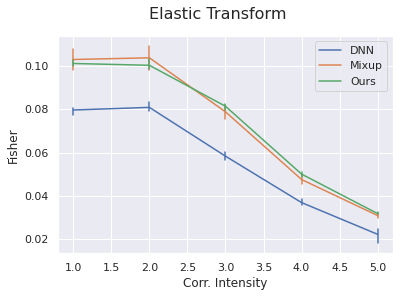}
     \end{subfigure}
     \hfill
     \begin{subfigure}[b]{0.3\textwidth}
         \centering
         \includegraphics[height=0.15\textheight]{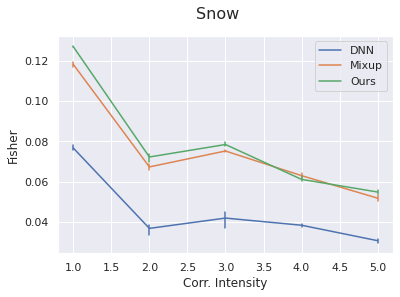}
     \end{subfigure} \\
     \begin{subfigure}[b]{0.3\textwidth}
         \centering
         \includegraphics[height=0.15\textheight]{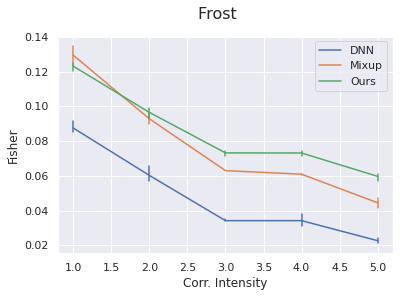}
     \end{subfigure}
     \hfill
     \begin{subfigure}[b]{0.3\textwidth}
         \centering
         \includegraphics[height=0.15\textheight]{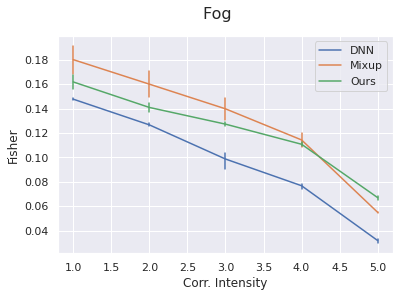}
     \end{subfigure}
     \hfill
     \begin{subfigure}[b]{0.3\textwidth}
         \centering
         \includegraphics[height=0.15\textheight]{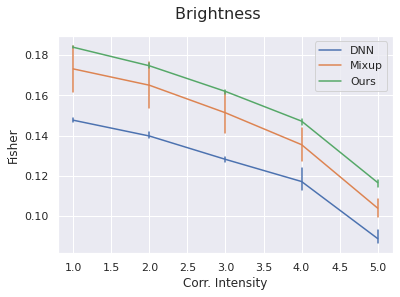}
     \end{subfigure} \\
     \begin{subfigure}[b]{0.3\textwidth}
         \centering
         \includegraphics[height=0.15\textheight]{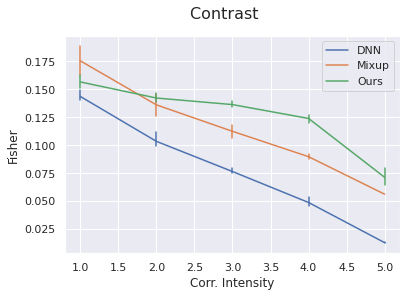}
     \end{subfigure}
     \hfill
     \begin{subfigure}[b]{0.3\textwidth}
         \centering
         \includegraphics[height=0.15\textheight]{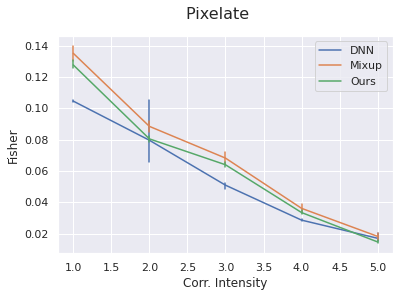}
     \end{subfigure}
     \hfill
     \begin{subfigure}[b]{0.3\textwidth}
         \centering
         \includegraphics[height=0.15\textheight]{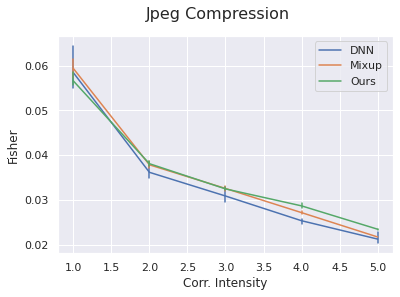}
     \end{subfigure}
        \caption{Fisher criterion for all the corruptions and intensity values of CIFAR-10-C (WRN28-10).}
        \label{fig:fisher_fig}
\end{figure}

Note, $\alpha$ would be high when within-class covariance is small and between-class covariance is high, thus, \textit{a high value of $\alpha$ is desirable}. In Figure~\ref{fig:fisher_fig}, we compute $\alpha$ over the C10 dataset with varying degrees of domain-shift. As the amount of corruption increases, $\alpha$ gradually decreases for all the models. However, \textit{RegMixup consistently provides the best $\alpha$} in most cases. 

\newpage 
\begin{table}
\resizebox{\linewidth}{!}
{
	\begin{tabular}{c|cc|cc}
		\toprule

  & \multicolumn{2}{c}{\textbf{WRN28-10}} & \multicolumn{2}{c}{\textbf{RN50}} \\
		& \multicolumn{1}{c}{\textbf{CIFAR10 (Val)}} & \multicolumn{1}{c}{\textbf{CIFAR100 (Val)}} & \multicolumn{1}{c}{\textbf{CIFAR10 (Val)}} & \multicolumn{1}{c}{\textbf{CIFAR100 (Val)}} \\
\midrule 
	\textbf{$\alpha$}	& \multicolumn{4}{c}{\textbf{Accuracy}}  \\
		\midrule
0.1 & 96.06 & 81.04 & 95.35 & 79.60  \\
0.2 & 96.46 & 80.91 & 95.21 & 80.11\\
0.3 & 96.77 & 81.06 & 95.36 & 80.31\\
0.4 & 96.71 & 81.01 & 95.26 & 78.93\\
0.5 & 96.70 & 80.99 & 95.28 & 78.91\\

1  & 96.74 & 80.66 & 94.96 & 78.79\\ 
5 & 96.62 & 79.84 & 94.98 & 77.74\\ 
10 & 96.54 & 79.24 & 94.94 & 75.76\\ 
20 & 96.26 & 78.40 & 95.16 & 75.56\\ 

\bottomrule 
\end{tabular}%

\hspace{1em}
\resizebox{0.2\linewidth}{!}
{
	\begin{tabular}{ccccccccccc|cccc}
		\toprule

& \textbf{RN50} \\ 
& \textbf{ImageNet} \\ 
\midrule 
$\alpha$ & \textbf{Accuracy} \\ 
		\midrule
0.1 & 77.10\\
0.2 & 77.02 \\ 
1  &76.19 \\
10 & 72.17 \\
20  & 71.51 \\
\bottomrule 
\end{tabular}%


}

}
\caption{Mixup hyperparameter cross-validation: Accuracy (\%) of WideResNet28-10 and ResNet50 on validation split of C10 and C100 for varying $\alpha$ values (table on the left), and of ResNet50 on ImageNet (table on the right).}
	\label{tab:bad_mixup}
\end{table}

\begin{table}
\centering
\resizebox{0.8\linewidth}{!}
{
	\begin{tabular}{c|cc|cc}
		\toprule

  & \multicolumn{2}{c}{\textbf{DenseNet-121}} & \multicolumn{2}{c}{\textbf{PyramidNet-200}} \\
		& \multicolumn{1}{c}{\textbf{CIFAR10 (Val)}} & \multicolumn{1}{c}{\textbf{CIFAR100 (Val)}} & \multicolumn{1}{c}{\textbf{CIFAR10 (Val)}} & \multicolumn{1}{c}{\textbf{CIFAR100 (Val)}} \\
\midrule 
	\textbf{$\alpha$}	& \multicolumn{4}{c}{\textbf{Accuracy}}  \\
		\midrule
0.1	&95.89 & 80.54 &	96.71 &	82.34 \\
0.2	&96.10 & 80.80 &	96.70 &	82.17 \\
0.3	&96.21 & 80.80 &	96.67 &	81.70 \\
0.4	&96.06 & 79.71 &	96.79 &	82.62 \\
0.5	&95.98 & 80.17 &	96.92 &	81.90 \\
1	&96.07 & 79.08 &	96.89 &	81.80 \\
10	&95.93 & 75.76 &	96.69 &	79.50 \\
20	&95.74 & 76.03 &	96.60 &	78.75 \\

\bottomrule 
\end{tabular}%

}
\caption{Mixup hyperparameter cross-validation: Accuracy (\%) of DenseNet-121 and PyramidNet-200. }
	\label{tab:pyr_dens}
\end{table}

\begin{table}
\centering
\resizebox{\linewidth}{!}
{
	\begin{tabular}{c|ccc|ccc|ccc|ccc}
		\toprule

  & \multicolumn{6}{c}{\textbf{WRN28-10}} & \multicolumn{6}{c}{\textbf{RN50}} \\
		& \multicolumn{3}{c}{\textbf{CIFAR10 (Val)}} & \multicolumn{3}{c}{\textbf{CIFAR100 (Val)}} & \multicolumn{3}{c}{\textbf{CIFAR10 (Val)}} & \multicolumn{3}{c}{\textbf{CIFAR100 (Val)}} \\
\midrule 
		 & \multicolumn{12}{c}{\textbf{Accuracy}}  \\
		\midrule
		\textbf{$\alpha$/$\eta$} & 0.1 & 1 & 2 & 0.1 & 1 & 2 &  0.1 & 1 & 2 & 0.1 & 1 & 2 \\
		\midrule 
0.1     & 96.12 & 96.02 & 96.38 & 80.08 & 80.86 & 80.96 & 94.32 & 95.02 & 94.48 & 78.56 & 79.84 & 77.92 \\ 
0.2     & 96.14 & 96.00 & 96.82 & 80.60 & 81.96 & 81.22 & 94.84 & 95.44 & 95.26 & 78.48 & 79.18 & 78.12 \\ 
0.3     & 95.82 & 96.68 & 96.34 & 80.92 & 81.82 & 80.94 & 94.62 & 95.46 & 95.48 & 78.38 & 79.74 & 78.21 \\ 
0.4     & 96.28 & 96.58 & 96.48 & 80.82 & 81.62 & 81.00 & 94.92 & 95.74 & 95.46 & 78.32 & 79.85 & 78.17 \\ 
0.5     & 96.08 & 96.88 & 96.44 & 81.00 & 81.20 & 81.38 & 94.72 & 96.04 & 95.24 & 78,56 & 79.34 & 78.29 \\ 
1       & 96.36 & 97.00 & 96.96 & 81.58 & 81.72 & 80.68 & 95.36 & 95.98 & 96.00 & 79.14 & 79.88 & 78.62 \\ 
5       & 96.50 & 97.14 & 97.22 & 82.00 & 81.94 & 81.04 & 95.86 & 96.26 & 96.04 & 79.42 & 80.13 & 78.30 \\ 
10      & 96.54 & 97.16 & 97.28 & 80.98 & 82.51 & 80.60 & 95.28 & 95.58 & 96.32 & 80.36 & 80.65 & 79.16 \\ 

15      & 96.65 & 97.27 & 97.18 & 81.38 & 82.16 & 80.59 & 95.15 & 96.10 & 96.34 & 79.72 & 80.32 & 78.80 \\ 

20      & 96.72 & 97.32 & 97.16 & 81.58 & 82.27 & 80.62 & 95.46 & 96.50 & 96.38 & 79.68 & 80.18 & 78.76 \\ 
30      & 96.74 & 97.28 & 97.28 & 82.48 & 81.94 & 80.40 & 95.66 & 96.46 & 95.90 & 79.39 & 79.78 & 79.22 \\

\bottomrule 
\end{tabular}%


}
\caption{RegMixup hyperparameter cross-validation: Accuracy (\%) of WideResNet28-10 and ResNet50 on validation split of C10 and C100 for varying $\alpha$ and $\eta$ values. }
	\label{tab:cv_regmixup}
\end{table}

\section{ECE results}
In this section we report the Expected Calibration Error (ECE) values. While ECE is more popular than AdaECE, the latter uses an adaptive binning scheme that better accounts for the bias introduced by the fact neural networks tend to concentrate most samples in high-confidence bins \citep{mukhoti2020calibrating}. 

\label{sec:ece}
\begin{table}
\resizebox{\linewidth}{!}
{
	\begin{tabular}{ccc|cc}
		\toprule
		& \multicolumn{4}{c}{\textbf{IND }} \\

& \multicolumn{2}{c|}{\textbf{WRN28-10}} & \multicolumn{2}{c}{\textbf{RN50}} \\
		& \multicolumn{1}{c}{\textbf{C10 (Test)}} & \multicolumn{1}{c|}{\textbf{C100 (Test)}} & \multicolumn{1}{c}{\textbf{C10 (Test)}} & \multicolumn{1}{c}{\textbf{C100 (Test)}}  \\

		\textbf{Methods} & \multicolumn{2}{c|}{\textbf{ECE ($\downarrow$)}} & \multicolumn{2}{c}{\textbf{ECE ($\downarrow$)}} \\
		\midrule

DNN               & 1.26  & 3.88 & 1.38   & 3.05 \\
Mixup           & 0.94   & 1.16 & \textbf{\underline{ 0.59}}    & 7.49     \\
RegMixup (\textbf{our})  & \textbf{\underline{0.62} }   & \textbf{\underline{1.65} } & 0.62  & \textbf{\underline{1.51}} \\
\cmidrule(r){2-5}                                                                                                
SNGP            & 0.84   & 1.95 & - & - \\
Augmix & 1.67  & 5.54 & - & - \\ 
\midrule
DE (5$\times$)  & 0.81  & 3.31 & 1.27  & 3.15 \\
\bottomrule 
\end{tabular}%

}
\caption{CIFAR IND calibration performance (\%).}
	\label{tab:onece_indc10c100}
      
\resizebox{\linewidth}{!}{

\begin{tabular}{cc|cccc}
	\toprule
	 & \multicolumn{1}{c|}{\textbf{IND}}  & \multicolumn{4}{c}{\textbf{Covariate Shift}}  \\

	& \multicolumn{1}{c|}{\textbf{ImageNet-1K (Test)}} & \multicolumn{1}{c|}{\textbf{ImageNet-R}} &
	\multicolumn{1}{c|}{\textbf{ImageNet-A}} &
	\multicolumn{1}{c|}{\textbf{ImageNet-V2}} &
	 \multicolumn{1}{c}{\textbf{ImageNet-Sk}}   \\

	  & \textbf{ECE} ($\downarrow$)  & \textbf{ECE} ($\downarrow$) & \textbf{ECE} ($\downarrow$) & \textbf{ECE} ($\downarrow$) & \textbf{ECE} ($\downarrow$)    \\

\midrule 

DNN           & 1.85 & 13.52 & 44.91 & 4.17 & 14.51 \\ 
Mixup     & \underline{\textbf{1.34}} & 12.13 & 44.65 & 4.34 & 15.32 \\
RegMixup (\textbf{our})  & 1.38 & 13.32 & \underline{\textbf{41.23}} & \underline{\textbf{3.41}} & 15.37 \\ 
\cmidrule(r){2-6}                                                                                                

AugMix    & 2.08 & \underline{\textbf{11.31}} & 42.85 & 3.97 & \underline{\textbf{14.32}} \\ 
\midrule
DE (5$\times$) & 1.39 & 13.59 & 42.92 & 4.07 & 17.34 \\

\bottomrule
	\end{tabular}%
}
    \caption{ImageNet calibration performance ($\%$).}
    \label{tab:imagenet_onece}

\resizebox{\linewidth}{!}
{
	\begin{tabular}{ccccc|cccc}
		\toprule
		& \multicolumn{8}{c}{\textbf{Covariate Shift}} \\
& \multicolumn{4}{c|}{\textbf{WRN28-10}} &  \multicolumn{4}{c}{\textbf{R50}} \\ 
		& \multicolumn{1}{c}{\textbf{C10-C}} &  \multicolumn{1}{c}{\textbf{C10.1}} &  \multicolumn{1}{c}{\textbf{C10.2}} &  \multicolumn{1}{c|}{\textbf{C100-C}} & \multicolumn{1}{c}{\textbf{C10-C}} &  \multicolumn{1}{c}{\textbf{C10.1}} &  \multicolumn{1}{c}{\textbf{C10.2}} &  \multicolumn{1}{c}{\textbf{C100-C}} \\

		\textbf{Methods} & \multicolumn{4}{c|}{\textbf{ECE ($\downarrow$)}} & \multicolumn{4}{c}{\textbf{ECE ($\downarrow$)}}  \\
		\midrule

DNN              & 12.64  & 4.29  & 9.03  & 9.96   & 12.31  & 4.53 & 8.97 & 19.80 \\
Mixup           & 7.54 & 4.28   & 7.40 & 10.32 & \textbf{\underline{10.23}} & 5.59 & 10.59 & 13.57  \\
RegMixup (\textbf{our})     & 9.56  & \textbf{\underline{2.87} } & \textbf{\underline{6.92} } & \textbf{\underline{7.98}}  & 12.61 & \textbf{\underline{3.25}} & \textbf{\underline{6.73}} & \textbf{12.87}      \\
\cmidrule{2-9}
SNGP             & 11.33   & 4.32   & 8.68 & 10.45 & - & - & - & -    \\
AugMix & \textbf{\underline{4.78}}  & 3.55  & 8.42  & 12.19 & - & - & - & - \\
\midrule 
DE (5 $\times$)   & 10.11  & 2.98  & 7.53  & 12.43   & 13.12 & 4.32 & 7.10 & \underline{12.53}      \\
\bottomrule 
\end{tabular}%

}
\caption{CIFAR CS calibration performance ($\%$).}
	\label{tab:onece_dshift}
\end{table}

\section{Using other vicinal distributions as regularizers}
\label{sec:additional}

In this Section we perform a preliminary analysis about using other techniques derived from Mixup as regularizers. In particular, we consider the following methods:
\begin{compactitem}
    \item CutMix \cite{CutMix}: instead of performing the convex combination of two samples, cuts and pastes a patch of an image into another, and the area of the patch is proportional to the interpolation factor $\lambda$. We set the CutMix probability to $p=1$ and cross-validate $\alpha \in \{0.1,0.2,0.3,1,10,20\}$, 
    \item RegCutMix: we replace Mixup with CutMix in our formulation. Same hyperparameters as above.
    \item Mixup + CutMix: we call this way the Transformers-inspired training procedure that at each iteration randomly decides (with equal probability) whether to apply Mixup or CutMix. For both Mixup and CutMix we cross-validate $\alpha \in \{0.1,0.3,1,10\}$.
    \item RegMixup + CutMix: as above, but using Mixup as a regularizer of the cross-entropy when Mixup is selected. Hyperparameters as above, $\eta \in \{0.1, 1, 3\}$ 
    \item RegMixup + RegCutMix: as above, but also using CutMix as a regularizer of the cross-entropy when CutMix is selected. Hyperparameters as above, $\eta$ is kept the same for both Mixup and CutMix.
\end{compactitem}
The results are reported in Tables \ref{tab:new_vicinals1} and \ref{tab:new_vicinals2}.

\begin{table*}
\resizebox{\linewidth}{!}
{
	\begin{tabular}{ccccc|ccc}
		\toprule
		& \multicolumn{1}{c}{\textbf{I.I.D. }} & \multicolumn{3}{c}{\textbf{Covariate Shift }} & \multicolumn{3}{c}{\textbf{O.O.D. }}\\
		& \multicolumn{1}{c}{\textbf{C10}} & \multicolumn{1}{c}{\textbf{C10-C}} &  \multicolumn{1}{c}{\textbf{C10.2}} &  \multicolumn{1}{c}{\textbf{C10.1}} & \multicolumn{1}{c}{\textbf{C100}} & \multicolumn{1}{c}{\textbf{SVHN}} & \multicolumn{1}{c}{\textbf{T-ImageNet}}\\

		\textbf{Methods} & \multicolumn{4}{c|}{\textbf{Accuracy ($\uparrow$)}} & \multicolumn{3}{c}{\textbf{AUROC ($\uparrow$)}}  \\
		\midrule

\textbf{WRN28-10} \\
\midrule 
 RegMixup  & \textbf{97.46} & \textbf{83.13} & 88.05 & 92.79 & 89.63 &\textbf{96.72} & 90.19 \\
  CutMix  & 96.70 & 71.40 & 87.06 & 91.18 &88.65& 92.25& \textbf{91.70}\\
 RegCutMix  & 96.79 & 71.26 & 86.84 & 92.10 &  89.24 &89.47& \textbf{91.68}\\
 Vit-Mixup+CutMix  & 97.23 & 77.60 & 87.85 & 92.30&  89.23 &  83.46 & 90.86 \\
ViT-RegMixup+CutMix& 97.30 & 78.67 & 88.42 & 93.15& 89.03 & 91.62 & 89.41 \\
ViT-RegMixup+RegCutMix& 97.47 & 79.10 & \textbf{88.79} & \textbf{93.35}& \textbf{89.93} & 94.18& 90.39\\
\midrule 

\textbf{R50} \\ 
\midrule 
RegMixup 	&96.71&	\textbf{81.18}	&86.72	&\textbf{91.58}&	\textbf{89.63}&	\textbf{95.39}	&\textbf{90.04} \\
CutMix  &96.27 & 70.63 & 85.35 & 90.35& 85.88 & 83.03 & 87.43 \\
 RegCutMix  &95.85 & 70.22 & 84.64 & 90.31&  85.10 & 84.76& 87.36 \\
 ViT-Mixup+CutMix & \textbf{96.87} &76.36 & \textbf{87.05} & 90.75& 89.19&  93.39 & 89.76 \\
ViT-RegMixup+CutMix &  96.60 & 77.21 & 86.30 & 90.97 & 84.22&87.90 & 85.87 \\
ViT-RegMixup+RegCutMix  & 96.47 & 75.36 & 86.01 & 91.28& 86.36 & 87.21 & 88.42 \\

\bottomrule
\end{tabular}%

}
\caption{Accuracy and out-of-distribution detection performance (\%) using other VRM techniques as regularisers for ResNet50 and WideResNet28-10 on CIFAR-10.}
	\label{tab:new_vicinals1}
\end{table*}

\begin{table*}
\resizebox{\linewidth}{!}
{
	\begin{tabular}{ccc|ccc}
		\toprule
		& \multicolumn{1}{c}{\textbf{I.I.D. }} & \multicolumn{1}{c}{\textbf{Covariate Shift }} & \multicolumn{3}{c}{\textbf{O.O.D. }} \\
		& \multicolumn{1}{c}{\textbf{C100}} & \multicolumn{1}{c}{\textbf{C100-C}}  & \multicolumn{1}{c}{\textbf{C10}} & \multicolumn{1}{c}{\textbf{SVHN}} & \multicolumn{1}{c}{\textbf{T-ImageNet}}\\

		\textbf{Methods} & \multicolumn{2}{c|}{\textbf{Accuracy ($\uparrow$)}} & \multicolumn{3}{c}{\textbf{AUROC ($\uparrow$)}}  \\
		\midrule

\textbf{WRN28-10} \\
\midrule 
RegMixup (Ours)&	83.25&	\textbf{59.44}&	\textbf{81.27}	&89.32&	\textbf{83.13} \\
CutMix  & 81.73 & 46.64 & 79.06 & 85.22 & 81.07 \\
RegCutMix & 82.30 & 47.37& 80.99&84.42& 81.85 \\
ViT-Mixup+CutMix & \textbf{84.05} & 54.94 & 80.81 & 85.36 & 81.32 \\
ViT-RegMixup+CutMix & 83.74 & 55.12& 79.92 & 86.44 & 82.28 \\
ViT-RegMixup+RegCutMix &  83.92 & 55.50& 80.75 & \textbf{90.23} & 83.07 \\

\midrule 
\textbf{R50} \\ 
\midrule 

RegMixup 	&81.52&	\textbf{57.64}&	79.44	&\textbf{88.66}	&82.56 \\
 CutMix  & 80.21 &45.23&  77.78 & 85.39& 80.33 \\
RegCutMix & 79.07 & 44.64&  77.56 &80.52& 79.18 \\
ViT-Mixup+CutMix& \textbf{82.39} & 56.40& 80.53 & 81.09 & 81.01 \\
ViT-RegMixup+CutMix&  81.65 & 52.99& 79.64 & 86.45& 81.42 \\
ViT-RegMixup+RegCutMix& 82.10 & 53.57& \textbf{80.66} & 84.43 & \textbf{83.03} \\
\bottomrule
\end{tabular}%

}
\caption{Accuracy and out-of-distribution detection performance (\%) using other VRM techniques as regularisers for ResNet50 and WideResNet28-10 on CIFAR-100.}
	\label{tab:new_vicinals2}
\end{table*}

As it can be seen: (1) in most cases RegCutMix underperforms with respect to CutMix, (2) in most cases RegMixup outperforms CutMix and RegCutMix, (3) Mixup+CutMix represents an extremely competitive method for in-distribution accuracy, but not as competitive for distribution shift and out-of-distribution detection, (4) RegMixup+CutMix is most of the times inferior to Mixup+CutMix, (5) RegMixup+RegCutMix is an extremely competitive method in several cases. We leave to future research exploring whether and how it is possible to combine Mixup-inspired techniques as regularizers to further improve the accuracy on i.i.d. and distribution-shifted data, and the out-of-distribution detection performance. We remark that in all these setups, RegMixup is extremely competitive and several times the best method, thus pointing out that despite its simplicity, it is extremely effective.

\end{document}